%% file: main.tex
\documentclass[10pt,twocolumn,letterpaper]{article}

\usepackage[pagenumbers]{cvpr} 

\usepackage{url}

\usepackage{color}
\usepackage{graphicx}
\usepackage{pifont}
\usepackage{multirow}
\usepackage{xspace}
\usepackage{caption}
\usepackage{amsmath}
\usepackage{enumitem}
\usepackage{wrapfig,lipsum}
\usepackage{diagbox}
\usepackage{makecell}
\usepackage{float}

\usepackage{booktabs}
\usepackage{siunitx}
\usepackage{dashrule}
\newcommand{\cmark}{\ding{51}\xspace}%
\newcommand{\xmark}{\ding{55}\xspace}%
\usepackage{colortbl}
\usepackage{csquotes}
\usepackage{enumitem}

\def\ourmethod{{\textit{OPAD}}\xspace}
\def\ourproblem{{\textit{1-SDP}}\xspace}

\def\ourste{{\textit{STE}}\xspace}

\newcommand{\minisection}[1]{\vspace{0.005in} \noindent {\bf #1}}

\newcommand{\tabincell}[2]{\begin{tabular}{@{}#1@{}}#2\end{tabular}}

\newcommand{\model}{\epsilon_\theta}

\newcommand{\conditioner}{\tau}
\newcommand{\expec}{\mathbb{E}}

\newcommand{\textprompt}{\mathcal{P}}
\newcommand{\textembedding}{\mathcal{C}}

\newcommand{\sdmodel}{\mathcal{G}}

\newcommand{\quotes}[1]{``#1''}

\usepackage{algorithm}
\usepackage{algpseudocode}

\usepackage[accsupp]{axessibility}  
\input{preamble}
\definecolor{cvprblue}{rgb}{0.21,0.49,0.74}
\usepackage[pagebackref,breaklinks,colorlinks,allcolors=cvprblue]{hyperref}

\def\paperID{***} 
\def\confName{CVPR}
\def\confYear{2026}

\title{Adversarial Concept Distillation for One-Step Diffusion Personalization}
\author{Yixiong Yang\textsuperscript{1,\thanks{Equal contribution.}}, 
Tao Wu\textsuperscript{2,3,\footnotemark[1]}, \\
Senmao Li\textsuperscript{4}, 
Shiqi Yang\textsuperscript{4,\thanks{Visiting researcher in Nankai University.}},   
Yaxing Wang\textsuperscript{4}, 
Joost van de Weijer\textsuperscript{2,3},
Kai Wang\textsuperscript{5,6,2,\thanks{Corresponding author}} 
\vspace{0.2cm}\\ 
\textsuperscript{1}Harbin Institute of Technology (Shenzhen), China \\
\textsuperscript{2}Computer Vision Center, Spain 
\textsuperscript{3}Universitat Autònoma de Barcelona, Spain \\
\textsuperscript{4}VCIP, CS, Nankai University, China \\
\textsuperscript{5}Program of Computer Science, City University of Hong Kong (Dongguan), China \\
\textsuperscript{6}City University of Hong Kong, HK SAR, China \\
}

\begin{document}
\maketitle
\input{sec/0_abstract}    
\input{sec/1_intro}

\input{sec/2_related_work}

\input{sec/3_method}

\input{sec/4_experiments}

\input{sec/5_conclusion}

\section*{Acknowledgments}
We acknowledge project PID2022-143257NB-I00, financed by MCIN/AEI/10.13039/501100011033 and ERDF/EU and FEDER, and the Generalitat de Catalunya CERCA Program, and ELLIOT project funded by the European Union under Grant Agreement 101214398.
This work was also supported by NSFC (NO. 62225604) and Youth Foundation (62202243).
We acknowledge \quotes{Science and Technology Yongjiang 2035} key technology breakthrough plan project and Chinese government-guided local science and technology development fund projects (scientific and technological achievement transfer and transformation projects) (254Z0102G).
Kai Wang acknowledges the funding from Guangdong and Hong Kong Universities 1+1+1 Joint Research Collaboration Scheme and the start-up grant B01040000108 from CityU-DG.

{
    \small
    \bibliographystyle{ieeenat_fullname}
    \bibliography{longstrings,main}
}

\input{sec/6_suppl}

\end{document}

%% file: sec/0_abstract.tex
\begin{abstract}

Recent progress in accelerating text-to-image diffusion models enables high-fidelity synthesis within a single denoising step. However, customizing the fast one-step models remains challenging, as existing methods consistently fail to produce acceptable results, underscoring the need for new methodologies to personalize one-step models. Therefore, we propose One-step Personalized Adversarial Distillation (OPAD), a framework that combines teacher–student distillation with adversarial supervision. A multi-step diffusion model serves as the teacher, while a one-step student model is jointly trained with it. The student learns from alignment losses that preserve consistency with the teacher and from adversarial losses that align its output with real image distributions. Beyond one-step personalization, we further observe that the student’s efficient generation and adversarially enriched representations provide valuable feedback to improve the teacher model, forming a collaborative learning stage. Extensive experiments demonstrate that OPAD is the first approach to deliver reliable, high-quality personalization for one-step diffusion models; in contrast, prior methods largely fail and produce severe failure cases, while OPAD preserves single-step efficiency. Code and data is available at \href{https://liulisixin.github.io/OPAD/}{https://liulisixin.github.io/OPAD/}.

\end{abstract}

%% file: sec/1_intro.tex
\section{Introduction}
\label{sec:intro}

Recently, large-scale generative models~\citep{ma2023overview_video_coding,tu2024overview_large_AI,xing2024survey_vdm,zhang2023text2image_survey} 
dominate high-quality text-to-image (T2I) generation and have been widely applied in diverse downstream tasks~\citep{hertz2022prompt,mou2023t2i,kai2023DPL,zhang2023controlnet,liu2025onepromptonestory,hu2024token_merging_tome, Luo_2025_ICCV}.
Among these applications, \textit{personalized} text-to-image generation, also referred to as \textit{new concept learning}~\citep{chung2025finetune_var,kumari2023customdiffusion,wu2025proxy_tuning_ar}, has emerged as a particularly important task. It involves adapting a T2I model to recognize and synthesize a novel concept from user-provided reference images.
Recent T2I personalization methods~\citep{textual_inversion,kumari2023customdiffusion,ruiz2023dreambooth} generally adapt pretrained T2I models using few-shot reference images and bind the novel concept to a pseudo-token so that the adapted model can synthesize various renditions of the new concept guided by text prompts.
Despite their success, the adapted T2I models still face a notable limitation which lies in their slow inference speed.
To address inference inefficiency for T2I models, recent research has turned to distillation-based acceleration techniques~\citep{liu2024distilled_DD,sauer2023adversarial,zheng2024trajectory_tcd,xu2025show_o_turbo}.
These techniques have matured considerably in the context of T2I diffusion models~\citep{luo2023lcm_lora,salimans2022progressive,sauer2023adversarial}, and we build on them in this paper.
In general, training-based distillation methods aim to learn a fast student generator~\citep{luo2023latent,sauer2023adversarial,song2023consistency,zheng2024trajectory_tcd} from a multi-step T2I diffusion teacher model. 
Representative acceleration methods~\citep{dao2024swiftbrushv2,luo2023latent,sauer2023adversarial} achieve impressive acceleration by reducing the number of sampling steps to four or even fewer than one step for image generation.

\begin{figure*}[t]
  \centering
  \includegraphics[width=0.999\linewidth]{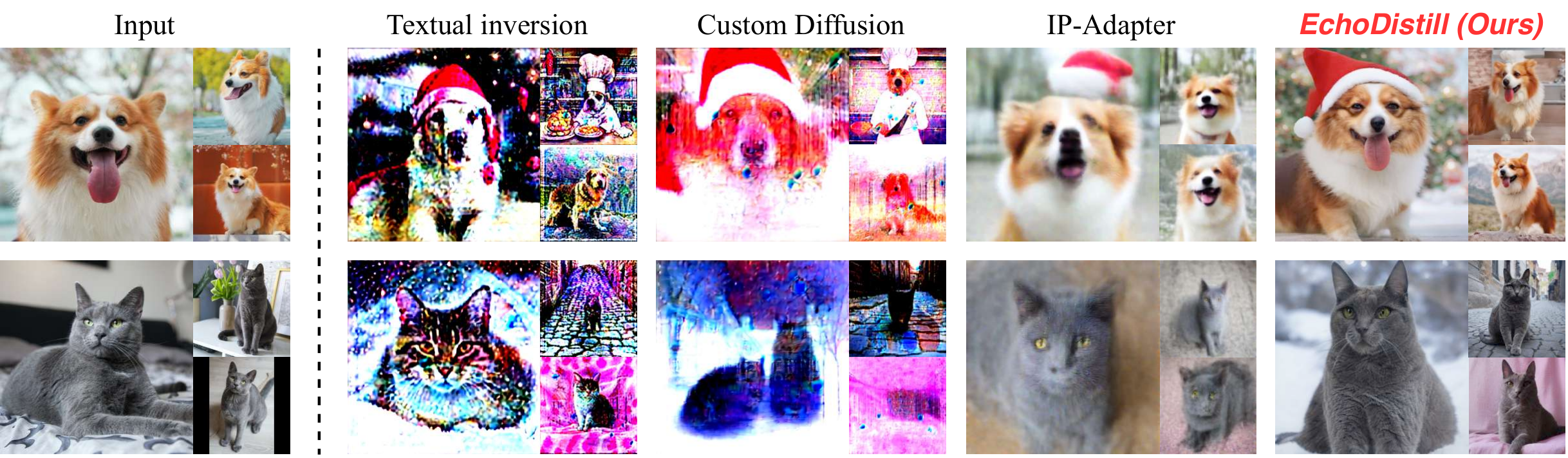}
  \caption{
  Comparison with existing new concept learning methods for one-step personalization: Textual Inversion~\citep{textual_inversion} and Custom Diffusion~\citep{kumari2023customdiffusion} for SDTurbo~\citep{sauer2023adversarial}, and IP-Adapter~\citep{ye2023ip-adapter} for \enquote{TCD~\citep{zheng2024trajectory_tcd}+SDXL~\citep{podell2023sdxl}}. 
  }  
  \label{fig:naive_baseline}
  \vspace{-2mm}
\end{figure*}

However, directly applying conventional personalization techniques to the fast few-step diffusion models results in severe failure cases. 
As the example illustrated in Fig.~\ref{fig:naive_baseline}, the \textit{word-inversion} method Textual Inversion applied to the one-step diffusion model SDTurbo~\citep{sauer2023adversarial} is unable to learn the textual concept tokens.
This leads to the first challenge in \ourproblem: 
\textbf{1) Student inadaptability:} The few-step student cannot learn text tokens independently and effectively.
This issue is further exemplified with the \textit{weight-optimization} method Custom Diffusion~\citep{kumari2023customdiffusion} (Fig.~\ref{fig:naive_baseline}), where jointly updating the text encoder and the diffusion backbone fails to improve performance and instead degrades generation quality. 
Note that existing \textit{encoder-based} personalization methods also struggle to generalize to one-step diffusion (see IP-Adapter in Fig.~\ref{fig:naive_baseline}).
A naive way to leverage the distillation technique for the \ourproblem problem is to first fine-tune a multi-step teacher model on the target concept, then use it to generate diverse samples to distill the one-step student. However, this results in two additional challenges:
\textbf{2) Inefficiency:} The multi-step generation process and non-end-to-end teacher-student distillation will significantly slow down learning.
\textbf{3) Teacher unreliability:} The teacher itself can also fail to capture certain concepts, limiting its effectiveness as a guiding signal for the student.
These issues contribute to significant failure cases of current methods in concept personalization for one-step diffusion models.

In this paper, we address the above challenges in \ourproblem by introducing the One-step Personalized Adversarial Distillation (\ourmethod) framework. 
Instead of relying on a sequential distillation pipeline, our proposed \ourmethod jointly optimizes a multi-step T2I teacher model and a one-step student model, allowing knowledge to be transferred effectively to the one-step setting. 
This design stabilizes knowledge transfer and leads to faster convergence. The student is guided by alignment losses to match the teacher’s output distribution, while adversarial supervision aligns its predictions with the real image distribution, resulting in faithful and realistic concept modeling. After one-step personalization, our method \ourmethod is also able to achieve few-step (2-step, 4-step, etc.) customized generations as a bonus. 
In addition, we introduce a collaborative learning stage tailored for the low-data nature of new concept learning. After acquiring the novel concept, the student model leverages its efficient one-step generation capability to synthesize additional samples of the concept. These synthetic samples serve as data augmentation for both models. We observe that further training on these enriched samples boosts generative performance for \textit{both} the student and the teacher, forming a mutually beneficial learning loop.
To summarize, we make the following contributions: 
\begin{itemize}[leftmargin=*]
    \item We are the \textit{first} to tackle the personalization problem on one-step diffusion models, termed \ourproblem, —a setting where existing techniques fail—achieving fast and faithful personalized concept generation. 
    \item To address \ourproblem, we propose solutions that include joint teacher–student training and a combination of alignment and adversarial guidance. We further propose a collaborative learning stage that enriches the limited concept data with student-generated samples, improving the generative capabilities of both the teacher and the student.
    \item Through extensive evaluations on DreamBench~\citep{ruiz2023dreambooth}, we show that prior methods severely break down under \ourproblem setting, whereas \ourmethod adapts reliably to new concepts while preserving high-quality one-step generation.
\end{itemize}

%% file: sec/2_related_work.tex
\section{Related works}

Text-to-image personalization, also known as new concept learning, focuses on adapting a model to a user-provided novel concept using a few reference images. 
This technique has been extensively studied in T2I diffusion models~\citep{butt2025colorpeel,textual_inversion,Cones2023,ruiz2023dreambooth} and recently studies in  the AR domain~\citep{chung2025finetune_var,wu2025proxy_tuning_ar}.

\minisection{Tuning-based methods}~\citep{li2025comprehensive_survey,zhao2025catversion,wang2024mcti} leverage reference images of the same concept to fine-tune either the T2I diffusion model or its learnable embeddings. Depending on the optimization targets, these methods can be categorized into word-inversion and weight-optimization approaches.

\textit{Word-inversion} methods focus on learning new concept tokens without modifying the model parameters. Textual Inversion~\citep{textual_inversion} is a pioneering approach that introduces pseudo-words by performing personalization in the text embedding space. 
Other works~\citep{agarwal2023image_matte,dong2022dreamartist,voynov2023ETI,tang2023iterinv,tang2024locinv} 
continually enable fine-grained and robust concept representation by employing designed loss functions to ensure that each token captures a distinct aspect of the reference images.
While these methods maintain high semantic consistency by keeping the generative model frozen, they suffer from limited identity fidelity due to the compression of rich image features into the low-dimensional text embedding space.

\textit{Weight-optimization} methods advance beyond token-level personalization by fine-tuning the model’s internal weights, enabling richer and more faithful concept learning. One of the most prominent methods is DreamBooth~\citep{ruiz2023dreambooth}, which fine-tunes a pre-trained text-to-image diffusion model to associate a unique identifier with a target subject using just 3–5 reference images. 
Following that, several methods such as Custom Diffusion (CD)~\citep{kumari2023customdiffusion} and Cones 2~\citep{Cones2023} propose optimizing only a subset of model parameters, significantly reducing both training time and memory consumption while preserving generation fidelity. Along similar lines, a variety of approaches~\citep{chen2023suti,gal2023e4t,han2023svdiff,zhang2022sine} have emerged to improve visual quality and efficiency. 
In addition to partial weight tuning, recent works~\citep{achlioptas2023stellar,xiang2023closer_look} introduce parameter-efficient strategies using Adapter modules, Low-Rank Adaptation (LoRA), or their variants, including Hyper-E4T~\citep{arar2023domain_agnostic}, DisenBooth~\citep{chen2023disenbooth}, etc.

\minisection{Tuning-Free methods}
have proposed encoder-based alternatives that significantly reduce or eliminate the need for fine-tuning backbones by leveraging pre-trained image encoders. These methods~\citep{li2023photomaker,rowles2024ipadapterinstruct,shi2023instantbooth,wang2024instantid,xiao2023fastcomposer,ye2023ip-adapter} enable efficient concept learning by extracting informative features from reference images using models trained on large-scale, diverse datasets. Some of them also specify in 
human face generation~\citep{cui2024idadapter,guo2024pulid,li2023photomaker,wang2024instantid,wu2024infinite_id}.
A recent advanced research is IP-Adapter~\citep{ye2023ip-adapter}, which utilizes the ViT image encoder from CLIP~\citep{radford2021clip} to extract reference image features. These features are then integrated into the diffusion model’s U-Net backbone through cross-attention mechanisms, resulting in more coherent and faithful renditions. 
While encoder-based methods are effective for general personalization from a single reference image, they are mainly tailored for large-scale, multi-step T2I models and require expensive retraining for each new backbone. Their limited ability to capture concept diversity~\citep{li2025comprehensive_survey} and lack of adaptation to few-step architectures remain key limitations. Applying such pretrained encoders or adapters to one-step diffusion models yields poor concept fidelity and image quality, highlighting a critical gap in current research.

%% file: sec/3_method.tex
\section{Methodology}
In this section, we first introduce the preliminaries in Sec.~\ref{sec:preliminary} and then present the key challenges of \ourproblem in Sec.~\ref{sec:challenge}. We finally present our framework, \ourmethod, to address these challenges in Sec.~\ref{sec:method}.

\subsection{Preliminaries}
\label{sec:preliminary}
\minisection{Latent Diffusion Models.}
LDM~\citep{Rombach_2022_CVPR_stablediffusion} is the most widely applied T2I diffusion models and the distillation teacher model for current few-step diffusion models~\citep{luo2023latent,sauer2023adversarial}. It is conditioned on textual input $\conditioner(\textprompt)$, where 
$\conditioner$ is the text encoder and $\textprompt$ is the prompt. 
The backbone $\model^{tc}$ is a conditional UNet~\citep{ronneberger2015unet} which predicts the added noise.
After predicting the noise, diverse schedulers~\citep{lu2022dpm,song2021ddim} are used to denoise.
Here, we use SD2.1 as the teacher model $\model^{tc}$ as in T2I acceleration approaches~\citep{nguyen2023swiftbrush,sauer2023adversarial}.

\minisection{One-Step Diffusion Model.}
To accelerate diffusion inference, various methods distill the sampling steps $T_{tc}=[1,T]$ of the teacher into few student anchor steps (NFEs\footnote{NFEs denote the number of function evaluations, from the view of diffusion ODE trajectories.}) $T_{st}=\left\{\upsilon_1,\ldots,\upsilon_n \right\}$ where $n$ is typically set to $1$, $2$, or $4$.
Specifically, a \textit{one-step} diffusion model $\sdmodel_{st}$ aims to transform a noise $x_T \sim \mathcal{N}(0, 1)$ directly into an image without iterative denoising steps, 
hence we denote this noise to image process as $x_0^{st} = \sdmodel^{st}_{\phi} (x_T,T,\textembedding)$. 
In this paper, we build on the one-step SD-Turbo~\citep{sauer2023adversarial} as the student model $\sdmodel^{st}_{\phi}$ for new concept learning under the \ourproblem setup.

\subsection{Core Challenges in \ourproblem}
\label{sec:challenge}

As previously illustrated qualitatively in Sec.~\ref{sec:intro} and supported quantitatively in Sec.~\ref{sec:expr}, conventional T2I personalization methods fail to learn new concepts in one-step diffusion models (such as SDTurbo~\citep{sauer2023adversarial}). We identify three main challenges in this \ourproblem setting.

\minisection{Student Inadaptability.}
We begin by applying the \textit{word-inversion} Textual Inversion~\citep{textual_inversion} to the one-step diffusion model SD-Turbo\citep{sauer2023adversarial}.
Observing from Fig.~\ref{fig:naive_baseline} and Table~\ref{tab:main_quantitative_results}, this naive adaptation fails to capture or reproduce the target concept, revealing a key limitation: one-step diffusion models cannot be effectively personalized by text encoder tuning alone, indicating the need for additional supervision or architectural changes.
We further evaluate \textit{weight-optimization} methods Custom Diffusion on SD-Turbo. This not only fails to enhance performance but also degrades image quality and concept fidelity, as proven in Fig.~\ref{fig:naive_baseline}. These findings suggest that excessive flexibility in updating the backbone of few-step models may disrupt the generative prior.
We hypothesize that this limitation arises from the inherent differences in the distillation objectives. Traditional diffusion models distill the entire generative process, preserving detailed noise-to-image mappings. In contrast, few-step models are typically trained using distribution alignment losses~\citep{poole2023dreamfusion,wang2023prolificdreamer}
rather than reconstructing individual denoising trajectories. 
As a result, applying conventional diffusion losses during personalization leads to ineffective learning and poor visual fidelity. 

\begin{figure*}[!t]
    \centering
    \includegraphics[width=0.99\linewidth]{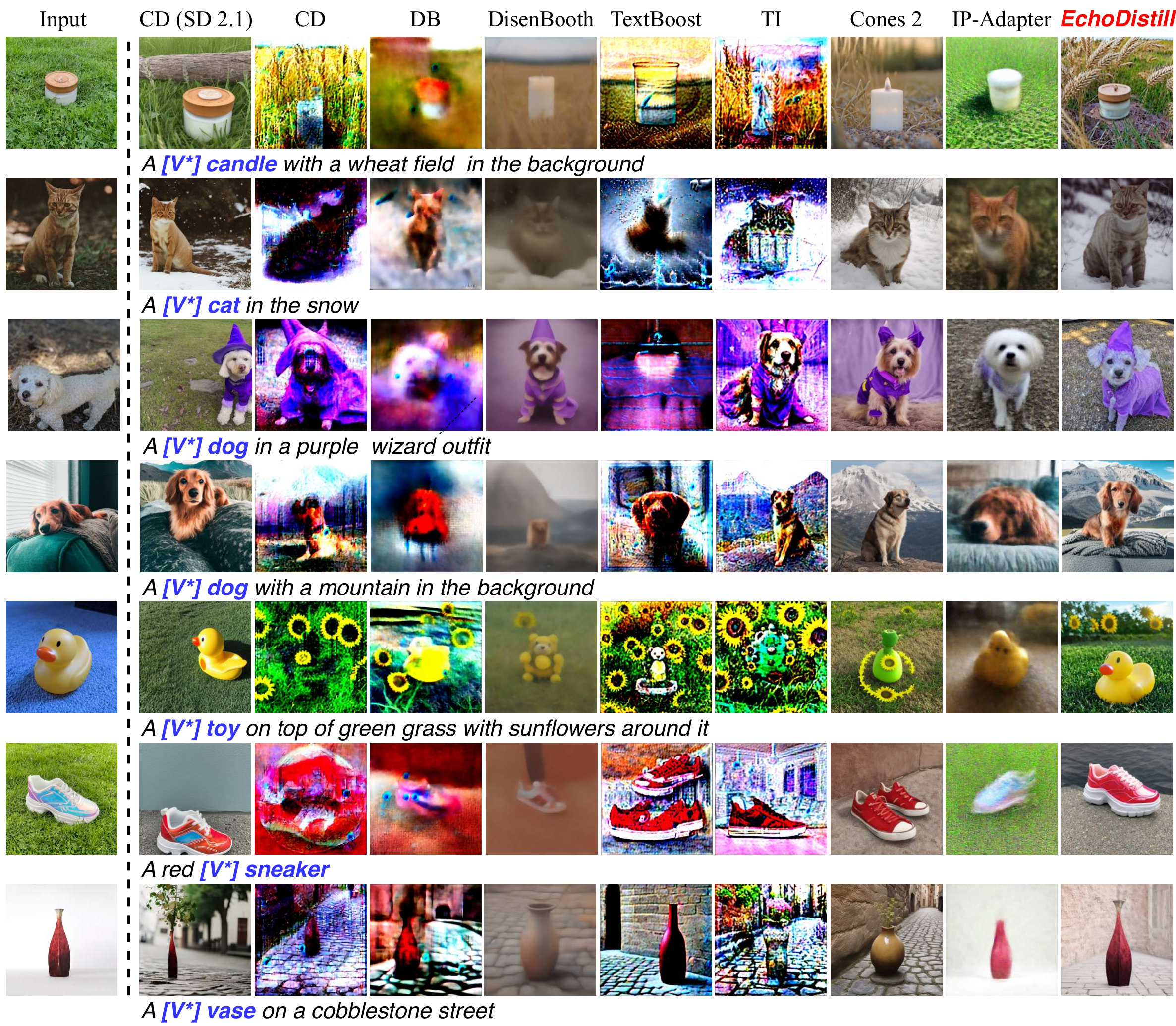}
    \vspace{-2mm}
    \caption{Our method \ourmethod (last column) compared with existing methods applied to the \ourproblem setup with SDTurbo~\citep{sauer2023adversarial} as the one-step diffusion backbone. One representive concept image is shown on the left-most column.
    }
    \vspace{-5mm}
    \label{fig:comparison}
\end{figure*}

\minisection{Teacher Irreliability and Inefficiency.}
A naive approach to tackle the \ourproblem problem is a two-stage distillation strategy, which we refer to as the \textit{teacher-first} distillation paradigm. In this setup, a multi-step teacher model is first fine-tuned on the target concept, and then used to generate diverse supervision samples for training the one-step student model. However, this paradigm faces two fundamental limitations:
\textit{1) Inefficiency:} The pipeline requires the teacher to first complete concept learning before it can supervise the student. Furthermore, the supervision involves multi-step generation (e.g., using 25 or 50 NFEs), making the process slow and computationally expensive.
\textit{2) Teacher Irreliability:} The teacher model may struggle to accurately learn certain visual concepts, particularly under sparse supervision, as can be seen from the Custom Diffusion baseline (SD2.1) in Table~\ref{tab:main_quantitative_results} and Fig.~\ref{fig:comparison}. When this occurs, the generated samples used for distillation are suboptimal or even misleading, thereby degrading the student model’s performance due to poor supervision.
Results for the teacher-first paradigm and additional discussions are provided in Sec.~\ref{appendix:Additional discussions for alternative designs} of the \textit{Appendix}.

\begin{figure*}[t]
    \centering
    \includegraphics[width=0.8\linewidth]{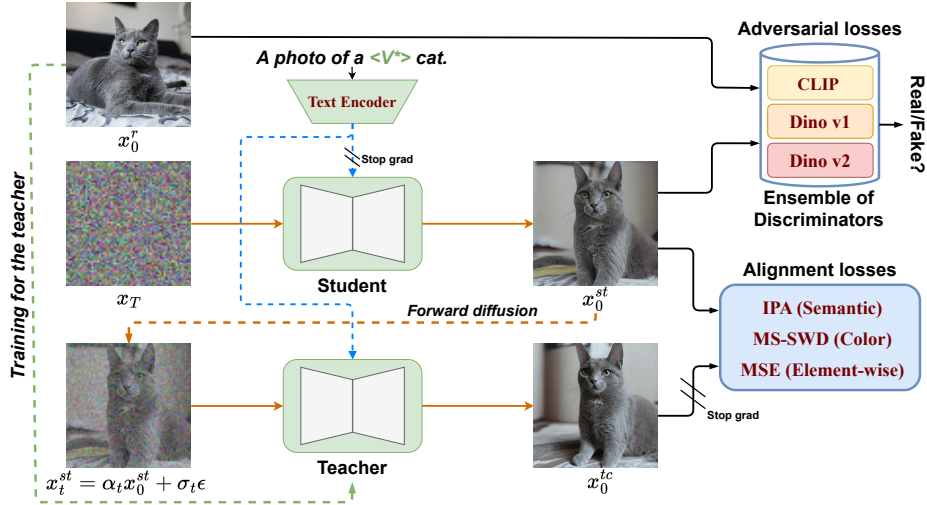}
    \vspace{-2mm}
    \caption{Overview of \ourmethod. The student and teacher jointly learn the new concept with a shared text encoder. The teacher learns from real images $x_0^r$ (\textcolor[rgb]{0.51,0.70,0.40}{green} line), and the text encoder is updated accordingly. The student is optimized with two objectives (\textcolor[rgb]{0.835, 0.42, 0.0}{gold} line): an adversarial loss to match real data distribution and alignment losses to match the denoised outputs of the teacher. The discriminators are trained to distinguish between the student's outputs and real images.
    }
    \label{fig:method}
    \vspace{-2mm}
\end{figure*}

\subsection{Our method}
\label{sec:method}

Based on the above observations, we propose our method One-step Personalization Adversarial Distillation (\ourmethod), as illustrated in Fig.~\ref{fig:method}. The overall framework is introduced in Sec.~\ref{subsubsec:overview}, and the loss formulation is detailed in Sec.~\ref{subsubsec:loss}. In Sec.~\ref{subsubsec:echoing}, we describe how student-generated samples are fed back into the training process. The whole workflow we refer to as \textit{Adversarial Concept Distillation}.

\subsubsection{Overview}
\label{subsubsec:overview}

 In OPAD, the student learns the new concept jointly with the multi-step teacher model. The teacher follows the standard way of Custom Diffusion~\citep{kumari2023customdiffusion}, while the student is supervised by alignment losses that encourage consistency with the teacher and by adversarial losses that align its outputs with the real image of the target concept.

Motivated by the benefits of consistent semantic grounding and the desire to simplify the training pipeline, we inherit the teacher’s text encoder for the student throughout training, forming a shared text encoder across both models. This design maintains a unified language–vision representation space and enables more reliable knowledge transfer.
To maintain memory efficiency and mitigate overfitting during personalization, we further adopt the lightweight adaptation strategy from Custom Diffusion~\citep{kumari2023customdiffusion}, updating only the key and value projection layers in both the teacher and student.

\ourmethod consists of three steps in each iteration. \textit{First}, the real image $x^r_{0}$ is fed into the teacher model. The \textit{teacher} is trained following the Custom Diffusion paradigm, where both the text encoder and the UNet are optimized using the noise prediction loss $\mathcal{L}_{\text{rec}}$:
\begin{equation}
    \mathcal{L}_{rec} = \expec_{x^r_{0}, y, t,  \epsilon \sim \mathcal{N}(0, 1)} 
    \Vert \epsilon - \model^{tc} (x_{t},t, \conditioner(\textprompt) \Vert_{2}^{2} 
    \label{eq:ldm_loss}
\end{equation}
\textit{Second}, the \textit{student} receives a random noise $x_T \sim \mathcal{N}(0, 1)$ input and generates an output $x_0^{\text{st}}=\sdmodel^{st} (x_T,T,\textembedding)$. 
This output is guided by a combination of two objectives: \textit{(1) alignment losses} between the student and teacher, and \textit{(2) adversarial losses} between the student and real images. For the alignment objective, $x_0^{\text{st}}$ is passed through the teacher's forward diffusion process to obtain a noisy version $x^{st}_t = \alpha_t x_0^{\text{st}} + \sigma_t \epsilon, \epsilon \sim \mathcal{N}(0,1)$, which is then denoised by the teacher to yield the predicted $x_0^{\text{tc}}$. 
This $x_0^{\text{tc}}$ is detached via a stop-gradient operation and serves as the supervision target for computing the alignment losses against $x_0^{\text{st}}$.
For the \textit{adversarial} objective, the student is optimized to fool an ensemble of discriminators, which are trained to distinguish the student-generated image $x_0^{\text{st}}$ from real images $x_0^r$.
\textit{Third}, the discriminators are optimized to enhance their discriminative performance.

\subsubsection{Distill with Alignment and Adversarial Guidance}
\label{subsubsec:loss}

Alignment losses supervise the student to match the teacher’s outputs, while adversarial losses encourage its predictions to align with the real image distribution. Our method is inspired by Sauer et al.~\cite{sauer2023adversarial} who  used adversarial losses to guide distillation in the context of training few-step models.
The detailed loss formulations are as follows.

\textbf{Alignment losses} encourage the student outputs to be semantically consistent with those from the teacher, capturing both low-level pixel details and high-level perceptual alignment. It is composed of three components: 
\begin{itemize}[leftmargin=*]
    \item Identity Feature Loss, adapted from IP-Adapter~\citep{ye2023ip-adapter} (IPA), extracts identity-preserving features from the image space $x_0$ using a CLIP image encoder followed by a projection network. Here the $x_0^{tc}$ is the estimated teacher image, computed from the teacher prediction $x_0^{tc} = \alpha^{-1/2}_t \cdot \{x_t - [ (1-\alpha_t)\cdot \epsilon_t^{tc}/(1-\bar{\alpha}_t)^{1/2}] \}$ while the $x_0^{st}$ is the image from student direct generation $\sdmodel^{st} (x_T,T,\textembedding)$. 
    This loss is computed as cosine similarity:
    \begin{equation}
    \label{eq:id_loss}
        \mathcal{L}_{\text{id}}(x_0^{st}, x_0^{tc}) = 1 - \cos \left(\text{IPA}(x_0^{st})), \text{IPA}(x_0^{tc})) \right)
    \end{equation}
    \item MSE Loss minimizes the distance between student and teacher latent representations:
    \begin{equation}
        \mathcal{L}_{\text{mse}}(x^{st}_0, x^{tc}_0) = \left\| x^{st}_0 - x^{tc}_0 \right\|_2^2
    \end{equation}
    
    \item Multi-Scale Sliced Wasserstein Distance~\citep{he2024multiscale}, compares multi-scale feature distributions in the image space to align structural and color information. This loss is proposed to alleviate the unstable color distribution during the distillation and defined as:
    \begin{equation}
        \mathcal{L}_{\text{swd}}(x_0^{st}, x_0^{tc}) = \text{MS-SWD}\left( x_0^{st}, x_0^{tc} \right)
    \end{equation}
\end{itemize}
The full alignment loss scales these components by weighting factors and a time-dependent term:
\vspace{-1mm}
\begin{equation}
\label{eq:align_loss}
\begin{aligned}
& \mathcal{L}_{\text{align}} = 
c(t) \cdot \big[\lambda_{\text{id}} \cdot \mathcal{L}_{\text{id}}(x_0^{st}, x_0^{tc}) \\
& + \lambda_{\text{mse}} \cdot \mathcal{L}_{\text{mse}}(x^{st}_0, x^{tc}_0) + \lambda_{\text{ms}} \cdot \mathcal{L}_{\text{swd}}(x_0^{st}, x_0^{tc})
\big]
\end{aligned}
\end{equation}

Inspired by Adversarial Diffusion Distillation (ADD)~\citep{sauer2023adversarial}, we introduce a timestep-dependent exponential weighting factor $c(t)=\alpha(t)$, where $t$ denotes the randomly sampled timestep in the teacher's noising–denoising process and $\alpha(t)$ is the same as defined in DDPM~\citep{ho2020ddpm}. At higher noise levels (i.e., larger $t$), the teacher’s predictions become increasingly unreliable, and the $c(t)$ is accordingly decreased. This design helps stabilize the student’s training by reducing the influence of noisy supervision.

\textbf{Adversarial losses} are designed to reduce the distribution gap between student-generated outputs and real concept images. Specifically, we ensemble multiple discriminators~\citep{chan2022eg3d,kumari2022ensembling}, each operating from a different semantic perspective, to improve training stability and achieve better results. We employ $K=3$ discriminators, each using a different pretrained backbone—DINOv1, DINOv2, and CLIP—as feature extractors. Each backbone is followed by a two-layer trainable projection head to distinguish between real and generated images, while the feature extractors remain frozen during training.
The adversarial loss for the student is defined as:
\begin{equation}
\label{eq:g_loss}
\mathcal{L}_{\text{GAN}}^{G} = \sum_{k=1}^{K} \lambda_k \cdot \mathbb{E}_{x_0^{st}} \left[ -\log(D_k(x_0^{st})) \right]
\end{equation}

where $D_k$ denotes the $k$-th discriminator (based on DINOv1~\citep{caron2021dino}, DINOv2~\citep{oquab2023dinov2} and CLIP~\citep{radford2021clip}), and $x_0^{st}$ is the student output image.
Denoting $x_0^r$ as the real image, the discriminator loss is defined as:
\begin{equation}
\begin{aligned}
\mathcal{L}_{\text{GAN}}^{D} = -\sum_{k=1}^{K} [ 
& \mathbb{E}_{x_0^r} [ \log D_k(x_0^r) ] \\
& + \mathbb{E}_{x_0^{st}} [ \log(1 - D_k(x_0^{st})) ] ]
\end{aligned}
\label{eq:d_loss}
\end{equation}

In summary, integrating a shared text encoder (\ourste), alignment losses and adversarial loss collectively enhances the adaptability and generalization capacity of the student model within the first distillation stage of our \ourmethod framework. 
\textit{Notably}, these loss formulations are specifically designed for the one-step student model, whose fast image generation allows efficient access to final outputs. In contrast, applying such losses to the multi-step teacher is \textit{impractical} due to the computational cost of obtaining real image outputs across \textit{iterative} denoising steps. A detailed algorithmic pipeline is given in Sec.~\ref{appendix:algorithm} of the \textit{Appendix}.

\subsubsection{Collaborative learning stage}
\label{subsubsec:echoing}
We interpret the one-step student model as a GAN-like generator, and it can benefit from aligning with the few-shot target \textit{data distribution} via \textit{adversarial training}. Specifically, incorporating the \textit{adversarial} loss~\citep{sauer2023adversarial,yin2024dmd2} as in Eq.~\ref{eq:g_loss} and Eq.~\ref{eq:d_loss} helps the student model generate samples that better match the \textit{distribution} of concept images, even enabling it to \textit{outperform the teacher model} in terms of \textit{visual qualities}, as demonstrated in Table~\ref{tab:main_quantitative_results}. 
This finding also aligns with insights from ADD~\citep{sauer2023adversarial}, which emphasizes the critical role of the discriminator loss in boosting generative \textit{fidelity}.

Building on this insight, we propose an additional collaborative learning stage, which leverages the student model’s rapid generation capability to provide constructive feedback to both itself and the teacher model. The collaborative learning stage mirrors the distillation stage, with the only difference being the definition of the real trainig examples $x_0^r$. 
Specifically, we replace real images with randomly generated samples from the updated student model after the first-stage distillation:  $\hat{x}_0^r = \sdmodel^{st} (x_T,T,\textembedding)$. The training objectives and update rules for both the student and teacher models remain unchanged from the initial distillation stage. The motivation behind this design is further discussed in Sec.~\ref{appendix:algorithm} of the supplementary material.

%% file: sec/4_experiments.tex
\section{Experiments}
\label{sec:expr}
\subsection{Experimental setups}

\begin{table*}[t]
\begin{minipage}[t]{0.55\linewidth}
\centering
\caption{Quantitative comparisons with existing personalization methods. }
\vspace{-2mm}
\resizebox{1.0\linewidth}{!}{
\begin{tabular}{c|c|
@{}c@{}|
@{}c@{}|
ccc}
\toprule
\textbf{Methods} & \textbf{Model} & \textbf{Train NFE} & \textbf{Infer NFE} & \textbf{CLIP-T} & \textbf{CLIP-I} & \textbf{DINO} \\
\midrule
\multirow{1}{*}{Custom Diffusion}  & SD 2.1  & 1000  & 25  & 0.269  & 0.752  & 0.519  \\

\midrule
\multirow{1}{*}{Custom Diffusion} & SD Turbo & 1       & 1    & 0.205  & 0.518  & 0.058    \\
Textual Inversion  & SD Turbo & 1 & 1 & \underline{0.252}  & 0.564  & 0.166   \\ 
Cones 2            & SD Turbo & 1 & 1 & \textbf{0.273}  & 0.619  & 0.204    \\
DreamBooth         & SD Turbo & 1 & 1 & 0.188  & 0.536  & 0.111  \\
TextBoost          & SD Turbo & 1 & 1 & 0.217  & 0.570  & 0.167  \\
DisenBooth         & SD Turbo & 1 & 1 & 0.251  & 0.564  & 0.231   \\
IP-Adapter         & SDXL+TCD      & 1 & 1 & 0.204  & \underline{0.628}  & \underline{0.325}  \\

\midrule 
\ourmethod  & SD Turbo & 1 & 1 &  \underline{0.252} & \textbf{0.783} & \textbf{0.637} \\
\bottomrule    
\end{tabular}
}
\label{tab:main_quantitative_results}
\end{minipage}
\hspace{1mm}
\begin{minipage}[t]{0.37\linewidth}
\centering
\caption{Ablation for components and infer-NFEs.}
\vspace{-2mm}
\resizebox{1.0\linewidth}{!}{
\begin{tabular}{ccccc}
\toprule
\multicolumn{2}{c}{\textbf{Methods}}  & \textbf{CLIP-T} & \textbf{CLIP-I} & \textbf{DINO} \\
\midrule
\multicolumn{2}{c}{\textbf{\textit{Ablate Components}}}   \\
\multicolumn{2}{c}{w/o teacher}        & 0.240 & 0.719 & 0.505 \\
\multicolumn{2}{c}{w/o discriminators} & 0.200 & 0.566 & 0.105 \\
\multicolumn{2}{c}{Full model}          & \textbf{0.252} & \textbf{0.783} & \textbf{0.637} \\
\midrule
\multicolumn{5}{l}{\textbf{\textit{Ablate Infer-NFEs}}}   \\
\multicolumn{2}{c}{1 step}  & 0.252       & \textbf{0.783}   & \textbf{0.637} \\
\multicolumn{2}{c}{2 steps} & 0.255       & \textbf{0.783}   & 0.635 \\
\multicolumn{2}{c}{4 steps} & \textbf{0.256}       & 0.772   & 0.610 \\
\bottomrule
\end{tabular}
}
\label{tab:ablation_components_and_steps}
\end{minipage}
\vspace{-3mm}
\end{table*}

\minisection{Comparison methods and evaluation metrics.}
We compare our method \ourmethod with the following T2I personalization approaches: 1) word-inversion: Textual Inversion~\citep{textual_inversion}, Cones 2~\citep{Cones2023}; 2) optimization-based: Custom Diffusion~\citep{kumari2023customdiffusion}, DreamBooth~\citep{ruiz2023dreambooth}, DisenBooth~\citep{chen2023disenbooth}, TextBoost~\citep{park2024textboost};
3) Encoder-based: IP-Adapter~\citep{ye2023ip-adapter}. 
For Disenbooth and Cones2 approach, the reference images were taken from training dataset in TextBoost.
For Custom Diffusion, we also include a baseline variant using SD2.1~\citep{Rombach_2022_CVPR_stablediffusion} as the backbone, which also serves as the base model for SDTurbo.
For IP-Adapter, as no publicly available implementation is compatible with the one-step SDTurbo model, we adopt an SDXL-based setup with TCD for comparison~\citep{podell2023sdxl, zheng2024trajectory_tcd}.
We follow the default configurations in their papers or open-source implementations. 
In the experiments, we evaluate our method on the DreamBooth~\citep{ruiz2023dreambooth} dataset, which contains 30 distinct concepts for personalized learning.
To measure the alignment between generated images and textual prompts, we employ the CLIP-T score.
Additionally, we assess cosine visual similarity between generated images and reference images using DINO~\citep{caron2021dino} and CLIP-I~\citep{radford2021clip} metrics, following standard practices in prior works~\citep{kumari2023customdiffusion,park2024textboost}.
More details of these comparisons and experiments are in Sec.~\ref{appendix:comparison} of the \textit{Appendix}.

\minisection{Implementation Details.}
During training of our method \ourmethod, the teacher model employs a sampling schedule with 1000 denoising steps (NFEs), while the student model performs denoising in a single step.
For the hyperparameters, the loss weights in Eq.~\ref{eq:align_loss} and Eq.~\ref{eq:g_loss} are set as $\lambda_{\text{ms}} = 0.1$, while all other weights are set to 1.0. The model is trained with a learning rate of $2 \times 10^{-5}$ and a batch size of 2.
All experiments are conducted on a single NVIDIA A40 GPU.

\subsection{Experimental results}

\minisection{Qualitative results.}
The main qualitative comparisons are presented in Fig.~\ref{fig:comparison}. All results were obtained by reproducing under the one-step personalization setup, except the multi-step Custom Diffusion method (2nd column). 
Among the seven baseline methods, Custom Diffusion, DreamBooth, TextBoost, and Textual Inversion fail to perform effective denoising or learn target concepts under the one-step inference setting.
DisenBooth and Cones2 struggle to capture precise concepts.
Although IP-Adapter preserves some identity consistency, its results are often blurry, misaligned with the prompts, and affected by the reference image background.
In contrast, our proposed method, \ourmethod, achieves both precise concept learning and strong semantic alignment between the generated images and input texts.

\minisection{Quantitative results.}
The detailed numeric results are presented in Table~\ref{tab:main_quantitative_results}. 
Notably, the first row reports the results of Custom Diffusion trained for 1,000 steps and evaluated with 25 inference steps. For fair comparison, all other methods, including ours, are evaluated under a setting where both the training and inference steps are set to 1.
\ourmethod maintains the text-image alignment quality competitive to the baselines, as evidenced by the CLIP-T Score. 
In terms of image similarity (CLIP-I, DINO scores), \ourmethod significantly outperforms the one-step or multi-step based methods. 
We \textit{note that} the CLIP-T is not that relevant in T2I personalization since it is computing the alignment with the generated image with the input text (without the conditional token), and therefore does not capture the alignment with the input concept images or the intended new concept.
Therefore the CLIP-I and DINO are more \textit{convincing} to demonstrate the effectiveness of personalization methods: on these scores our method shows the best performance.
Moreover, \ourmethod even surpasses the Custom Diffusion (SD2.1) in the first row, further supporting our argument that the teacher model is not fully reliable under the \ourproblem setup.
Additionally, the low CLIP-I and DINO scores of the Textual Inversion baseline highlight the student’s inability to independently learn the concept without effective supervision.
These results collectively validate the key challenges we identified in adapting concept learning to the \ourproblem setup.

\begin{figure*}[t]
    \centering
    \includegraphics[width=0.99\linewidth]{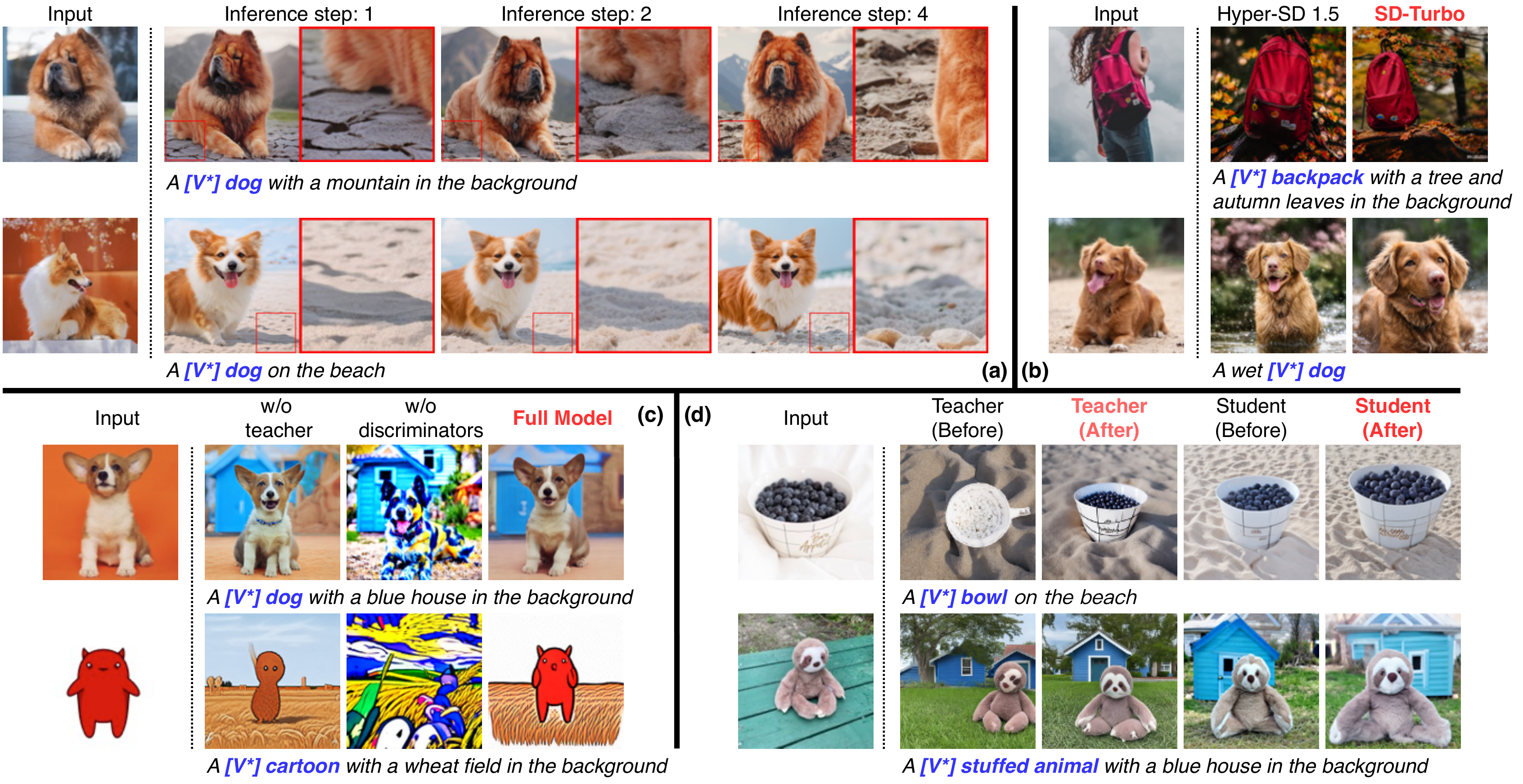}
    \vspace{-3mm}
    \caption{(a) Abating the Infer-NFEs; (b) Ablating the one-step diffusion backbone; (c) Ablating the teacher and discriminators; (d) Ablating the collaborative learning stage.
    }
    \label{fig:ablation_study}
    \vspace{-4mm}
\end{figure*}

\begin{table}[t]
\vspace{-1mm}
\centering
\caption{Ablations for backbones and collaborative learning stage.}
\vspace{-2mm}
\resizebox{0.78\linewidth}{!}{
\begin{tabular}{ccccc}
\toprule
\multicolumn{2}{c}{\textbf{Methods}} & \textbf{CLIP-T} & \textbf{CLIP-I} & \textbf{DINO} \\
\midrule
\multicolumn{2}{c}{\textbf{\textit{Ablate 1-Step Backbone}}}    \\
\multicolumn{2}{c}{Hyper-SD1.5}  & 0.211          & 0.709          & 0.463 \\
\multicolumn{2}{c}{SD-turbo}     & \textbf{0.252} & \textbf{0.783} & \textbf{0.637} \\

\midrule
\multicolumn{5}{l}{\textbf{\textit{Ablate  Collaborative Learning Stage}}}   \\
\multirow{2}{*}{Teacher} & Before	& \textbf{0.269}	& 	0.752	& 	0.519 \\
 & After		& 0.265	& 	\textbf{0.764}	& 	\textbf{0.571} \\
\cmidrule(r){3-5}
\multirow{2}{*}{Student} & Before	& 	\textbf{0.252}		& 0.783	& 	0.637 \\
 & After		& 0.236	& 	\textbf{0.798}	& 	\textbf{0.673} \\
\bottomrule
\end{tabular}
}
\label{tab:ablation_backbone_and_echo}
\vspace{-5mm}
\end{table}

\subsection{Ablation Study}
\label{Ablation Study}
\minisection{Main components.}
Table~\ref{tab:ablation_components_and_steps} presents the analysis of removing key components from \ourmethod, namely the teacher model and the discriminators. As shown, removing the teacher or the discriminators leads to a noticeable decline across all evaluation metrics. These results indicate that both components play a crucial role in effectively learning new concepts.
Qualitative ablation results are shown in Fig.~\ref{fig:ablation_study}-(c). Our full model generates high-quality and semantically consistent outputs. In contrast, omitting the teacher results in images that lack fine details specific to the target concept.
When the discriminators are removed, the outputs are significantly more noisy. This degradation likely stems from the teacher’s use of a single denoising step, which produces $\mathbf{x}_0$ predictions with residual noise that are subsequently propagated to the student model during training.

\minisection{Inference Steps.}
Although \ourmethod is trained for a 1-step setting, we further evaluate its performance for 2-step and 4-step denoising, using the same trained model without re-training. As reported in Table~\ref{tab:ablation_components_and_steps}, CLIP-T scores show slight improvements with additional steps, whereas CLIP-I and DINO scores exhibit marginal declines. Overall, the variations across quantitative metrics remain minimal. 
The qualitative comparisons in Fig.~\ref{fig:ablation_study}-(a) reveal more perceptible differences. Notably, increasing the number of inference steps enhances image fidelity, especially through richer background details and finer textures. This generalizability emerges as a beneficial byproduct of training \ourmethod.

\minisection{1-Step Backbones.}
We perform a backbone ablation study to assess the adaptability of our method to one-step backbones. In particular, we replace the student model with Hyper-SD1.5 and adjust the teacher model accordingly to SD1.5. The quantitative results are summarized in Table~\ref{tab:ablation_backbone_and_echo}.
Our findings indicate that this alternative backbone yields inferior performance compared to the SDTurbo backbone. However, as illustrated in Fig.\ref{fig:ablation_study}-(b), Hyper-SD1.5 still remains capable of generating reasonable outputs in such cases. Notably, our choice of SDTurbo as the primary backbone is motivated by two main factors: it is one of the few one-step models capable of generating high-quality images, and its distillation-based training process is well aligned with our framework, which likely contributes to its superior performance over Hyper-SD1.5 when used as the backbone.

\minisection{Collaborative Learning Stage.}
In Table~\ref{tab:ablation_backbone_and_echo}, we compare the student and teacher performance after the collaborative learning stage. The teacher model exhibits significant improvements in CLIP-I and DINO scores, while CLIP-T scores experience a slight decline. These results suggest that \textit{the student’s output can effectively enhance the performance of both teacher and student models, particularly in terms of identity and visual similarity.}
Qualitative examples in Fig.~\ref{fig:ablation_study}-(d) further support this \textbf{observation}: {when the teacher model struggles to learn certain concepts, leveraging the student’s output as an additional supervisory signal enables the teacher to better capture and reproduce those challenging concepts.}
This observation is also consistent with our theoretical analysis presented in \cref{subsubsec:echoing}.

Extended results are provided in~\cref{appendix:comparison}--\cref{CustomConcept101 Dataset} of the \textit{Appendix}, including alternative designs, user study, experiments on the CustomConcept101 dataset~\citep{kumari2023customdiffusion}, etc.

%% file: sec/5_conclusion.tex
\section{Conclusions}
In this paper, we introduced the novel task of one-step diffusion personalization (\ourproblem), a significant step toward bridging the gap between fast generative inference and concept-personalized image synthesis.
We identified three major challenges that prevent conventional personalization methods from being directly applicable to one-step diffusion models. 
To overcome these limitations, we proposed a unified framework, \ourmethod, to address them. \ourmethod integrates joint teacher–student training with distillation guided by alignment and adversarial supervision. We also introduce a collaborative learning stage that uses the outputs of the one-step student model to further enhance training. Through these components, \ourmethod enables effective adaptation to new visual concepts. 
Extensive experiments on the DreamBench benchmark confirm that \ourmethod consistently outperforms existing personalization approaches, setting a new foundation for rapid and reliable concept learning in diffusion-based generation.

%% file: sec/6_suppl.tex
\clearpage
\maketitlesupplementary

\section{Broader Impacts}

Our method, \ourmethod, enables efficient and high-fidelity personalized image generation, which holds significant potential for a range of creative applications, including design, education, and virtual content creation. By reducing the need for extensive data and computation, \ourmethod democratizes access to advanced generative tools, empowering users with minimal resources to produce customized visual content.
However, as with other powerful generative models, our approach also introduces potential risks. These include the unauthorized generation of content, impersonation, and the creation of misleading or harmful imagery. We acknowledge these risks and stress the importance of deploying appropriate safeguards, such as content moderation, usage auditing, and user authentication mechanisms, particularly in real-world applications.
We advocate for the responsible use of this technology and encourage the research community and stakeholders to collaborate on developing ethical guidelines and technical solutions to mitigate potential misuse.

\section{Limitations}
Our proposed method, \ourmethod, marks an initial step toward enabling one-step diffusion models to learn novel concepts efficiently. While the experimental results are promising, several limitations remain and warrant further investigation: 
\textit{(1) Training efficiency:} The current training pipeline is hindered by the computational overhead introduced by the multi-discriminator architecture. Optimizing or rethinking this component could significantly improve training speed.
\textit{(2) Limited one-shot personalization:} The discriminator’s reliance on multiple reference samples to model the underlying data distribution makes true one-shot personalization challenging. Designing a more robust discriminator or alternative mechanisms to enable faithful learning from a single image remains an open problem.
\textit{(3) Training instability:} As with many GAN-based methods, our approach may exhibit instability across runs, particularly for challenging concepts, where achieving optimal results may require a few trials. Enhancing training stability remains a promising direction for future work.
We leave these challenges as compelling avenues for future research, aiming to build upon this initial framework to support broader generalization, compositionality, and efficiency.

\section{Ethical and LLM Statements}
We acknowledge the potential ethical implications of deploying generative models, including concerns related to privacy, data misuse, and the propagation of biases. All models used in this paper are publicly available, and we will release the modified codes to enable reproduction of our results. 
We also emphasize the potential misuse of customization approaches in generating misinformation, and we strongly encourage and support their responsible usage. 
Regarding the use of LLMs, we clarify that in this work they were only minimally employed, specifically for correcting grammatical errors.

\section{Overview of T2I Personalization Methods}
\label{appendix:overview}
In this section, we present a comprehensive comparison of representative text-to-image personalization methods, expanding upon the overview introduced in $\lambda$-Eclipse~\citep{patel2024lambda_eclipse}. Table~\ref{tab:all_methods} provides an extended summary that systematically contrasts these approaches across several key dimensions, including support for single- or multi-subject personalization, training-free versus training-based paradigms, number of input images required, inference efficiency, etc.

\begin{table*}[!t]
\centering
\small
\caption{
We provide an overview of representative text-to-image personalization methods by extending the summary introduced in $\lambda$-Eclipse~\citep{patel2024lambda_eclipse}. The base models listed correspond to those used in their original papers. For a fair comparison with the highlighted methods in our study, we re-implemented and adapted all approaches using the same base model configuration as described in our main paper.
ChilloutMix is a community-contributed variant of the Stable Diffusion model~\citep{Rombach_2022_CVPR_stablediffusion}. Inifinity refers to a variant of the Text-to-Image VAR model~\citep{tian2024VAR}, while LlamaGen denotes a text-to-image auto-regressive (AR) model~\citep{sun2024llamagen}.
} 
\setlength{\tabcolsep}{8pt}
\resizebox{0.75\textwidth}{!}{
\begin{tabular}{r|cccccc}
\hline\toprule
\multirow{2}{*}{\textbf{\tabincell{c}{Method}}} & \multirow{2}{*}{\textbf{\tabincell{c}{Multi-\\Subject}}}&\multirow{2}{*}{\textbf{\tabincell{c}{Tuning-\\Free}}} &\multirow{2}{*}{\textbf{\tabincell{c}{Base\\Model}}} & \multirow{2}{*}{\textbf{\tabincell{c}{Input\\Images}}}  & \multirow{2}{*}{\textbf{\tabincell{c}{Inference\\Steps}}} & \multirow{2}{*}{\textbf{\tabincell{c}{Note}}} \\
& \\
\midrule
\rowcolor{cyan}
Textual Inversion~\citep{textual_inversion} &\xmark &\xmark &SDv1.4 & Few-Shot & Multi-Step & Word-Inversion\\
P+~\citep{voynov2023ETI} &\xmark &\xmark &SDv1.4 & Few-Shot & Multi-Step & Word-Inversion\\
ProsPect~\citep{zhang2023prospect} &\xmark &\xmark &SDv1.4 & 1-Shot & Multi-Step & Word-Inversion\\
MATTE~\citep{agarwal2023image_matte} &\xmark &\xmark &SDv1.4 & 1-Shot & Multi-Step & Word-Inversion\\
\rowcolor{cyan}
Cones 2~\citep{liu2023customizable_cones2} &\cmark &\xmark &SDv2.1 & Few-Shot & Multi-Step & Word-Inversion \\

\rowcolor{cyan}
DreamBooth~\citep{ruiz2023dreambooth} &\xmark &\xmark &SDv1.4 & Few-Shot & Multi-Step &\\
ClassDiffusion~\citep{huang2024classdiffusion} &\xmark &\xmark &SDv1.5 & Few-Shot & Multi-Step\\
\rowcolor{cyan}
DisenBooth~\citep{chen2023disenbooth} &\xmark &\xmark &SDv2.1 & 1-shot & Multi-Step &\\
CatVersion~\citep{zhao2025catversion} &\xmark &\xmark &SDv1.5 & Few-Shot & Multi-Step\\
AttnDreamBooth~\citep{pang2024attndreambooth} &\xmark &\xmark &SDv2.1 & 1-shot & Multi-Step\\

ViCo~\citep{tumanyan2023plug} &\xmark &\xmark &SDv1.4 & Few-Shot & Multi-Step\\
\rowcolor{cyan}
TextBoost~\citep{park2024textboost}  &\xmark &\xmark &SDv1.5 & 1-shot & Multi-Step &\\

NeTI~\citep{alaluf2023neural_neti} &\xmark &\xmark &SDv1.4 & Few-Shot & Multi-Step\\
HyperDreamBooth~\citep{ruiz2023hyperdreambooth} &\xmark &\xmark &SDv1.5 & 1-shot & Multi-Step\\

E4T~\citep{gal2023e4t} &\xmark &\xmark &SD & 1-shot & Multi-Step\\
Hyper-E4T~\citep{arar2023domain_agnostic} &\xmark &\xmark & SD & 1-shot & Multi-Step\\

ARBooth~\citep{chung2025finetune_var} &\xmark &\xmark & Infinity~\citep{han2024infinity} & Few-Shot & Multi-Step\\
Proxy-Tuning~\citep{wu2025proxy_tuning_ar} &\xmark &\xmark & LlamaGen~\citep{sun2024llamagen} & Few-Shot & Multi-Step\\

Continual Diffusion~\citep{smith2023continual} &\cmark &\xmark & SD & Few-Shot & Multi-Step\\
Perfusion~\citep{tewel2023keylocked_perfusion} &\cmark &\xmark &SDv1.5 & Few-Shot & Multi-Step\\
\rowcolor{cyan}
Custom Diffusion~\citep{kumari2023customdiffusion} &\cmark &\xmark &SDv1.4 & Few-Shot & Multi-Step &\\
Cones~\citep{Cones2023} &\cmark &\xmark &SDv1.4 & 1-shot & Multi-Step\\
SVDiff~\citep{han2023svdiff} &\cmark &\xmark &SDv1.5 & Few-Shot & Multi-Step\\
FreeCustom~\citep{ding2024freecustom} &\cmark &\xmark &SDv1.5 & 1-Shot & Multi-Step\\
Mix-of-Show~\citep{gu2024mixofshow} &\cmark &\xmark &Chilloutmix & Few-Shot & Multi-Step\\
LoRACLR~\citep{simsar2024loraclr} &\cmark &\xmark &Chilloutmix & Few-Shot & Multi-Step\\
Orthogonal~\citep{po2024orthogonal} &\cmark &\xmark &Chilloutmix & Few-Shot & Multi-Step\\
OMG~\citep{kong2024omg} &\cmark &\xmark & SDXL & Few-Shot & Multi-Step\\
Zip-LoRA~\citep{shah2023ziplora} &\cmark &\xmark &SDXL & Few-Shot & Multi-Step\\

Break-A-Scene~\citep{avrahami2023breakascene} &\cmark &\xmark &SDv2.1 & 1-shot & Multi-Step\\ 

TokenVerse~\citep{garibi2025tokenverse} &\cmark &\xmark & Flux~\citep{flux2024} & 1-shot & Multi-Step\\

\midrule
\rowcolor{green}
\ourmethod (Ours) &\xmark &\xmark & SDTurbo~\citep{sauer2023adversarial} & Few-Shot & 1-step &\\
\midrule
PhotoMaker~\citep{li2023photomaker} &\xmark &\cmark &SDXL & 1-shot & Multi-Step & Human Face \\
ConsistentID~\citep{huang2024consistentid} &\xmark &\cmark &SDv1.5 & 1-shot & Multi-Step &Human Face\\
InstantID~\citep{wang2024instantid} &\xmark &\cmark &SDXL & 1-shot & Multi-Step &Human Face\\ 
Profusion~\citep{zhou2023enhancing_profusion} &\xmark &\cmark &SDv2 & 1-shot & Multi-Step &Human Face\\
PuLID~\citep{guo2024pulid} &\xmark &\cmark &SDXL & 1-shot & Multi-Step &Human Face\\
Infinite-ID~\citep{wu2024infinite_id}&\xmark &\cmark &SDXL & 1-shot & Multi-Step &Human Face\\
LCM-Lookahead~\citep{gal2024lcm_lookahead}&\xmark &\cmark &SDXL & 1-shot & Multi-Step &Human Face\\
InfiniteYou~\citep{jiang2025infiniteyou} &\xmark &\cmark & Flux~\citep{flux2024} & 1-shot & Multi-Step &Human Face\\
\rowcolor{cyan}
IP-Adapter~\citep{ye2023ip-adapter} &\xmark &\cmark &SDv1.5 & 1-shot & Multi-Step &\\
ELITE~\citep{Wei2023ELITEEV} &\xmark &\cmark &SDv1.4 & 1-shot & Multi-Step\\
UMM-Diffusion~\citep{ma2023unified} &\xmark &\cmark &SDv1.5 & 1-shot & Multi-Step\\
InstantBooth~\citep{shi2023instantbooth} &\xmark &\cmark &SDv1.4 & Few-Shot & Multi-Step\\
BLIP-Diffusion~\citep{li2023blip} &\xmark &\cmark &SDv1.5 & 1-shot & Multi-Step\\
JeDi~\citep{zeng2024jedi} &\xmark &\cmark &SDv1.4 & Few-Shot & Multi-Step\\
Re-Imagen~\citep{chen2022re} &\xmark &\cmark & Imagen~\citep{saharia2022imagen} & 1-shot & Multi-Step\\
SuTi~\citep{chen2023suti} &\xmark &\cmark &Imagen~\citep{saharia2022imagen} & Few-Shot & Multi-Step\\
Taming~\citep{jia2023taming} &\xmark &\cmark &Imagen~\citep{saharia2022imagen} & 1-shot & Multi-Step\\

Kosmos-G~\citep{pan2023kosmos} &\cmark &\cmark &SDv1.5 & 1-shot & Multi-Step\\
SSR-Encoder~\citep{zhang2024ssr_encoder} &\cmark &\cmark &SDv1.5 & 1-shot & Multi-Step\\

$\lambda$-Eclipse~\citep{patel2024lambda_eclipse}&\cmark &\cmark & Kandinsky~\citep{arkhipkin2023kandinsky} & 1-shot & Multi-Step\\
FastComposer~\citep{xiao2023fastcomposer} &\cmark &\cmark &SDv1.5 & 1-shot & Multi-Step\\
Subject-Diffusion~\citep{ma2023subject_diffusion} &\cmark &\cmark &SDv2.1 & 1-shot & Multi-Step\\
RMCC~\citep{huang2024resolving_rmcc} &\cmark &\cmark & SDXL & 1-shot & Multi-Step\\
Emu2~\citep{sun2023generative_emu2} & \cmark & \cmark & SDXL & 1-shot & Multi-Step \\
MS-Diffusion~\citep{wang2024ms_diffusion} &\cmark &\cmark &SDXL & 1-shot & Multi-Step\\

   \bottomrule
   \hline
\end{tabular}
}
\label{tab:all_methods}
\end{table*}

\section{Algorithmic Description of \ourmethod Training}
\label{appendix:algorithm}
\minisection{Distillation Stage. }
The full training procedure of one-step personalization in \ourmethod is described in Sec.~\ref{sec:method} of the main paper. For completeness, Algorithm~\ref{alg:training} provides the detailed step-by-step implementation. Each iteration consists of three steps: 
\textit{(1)} the teacher model and text encoder are jointly optimized via the noise prediction loss $\mathcal{L}_{\text{rec}}$ following the Custom Diffusion paradigm; 
\textit{(2)} training of the student model through a combination of alignment losses with the teacher’s outputs and adversarial losses against real image data; 
and \textit{(3)} updating the discriminators to improve their capacity to differentiate between real and synthesized samples. 

\begin{algorithm*}[t]
\caption{Training Pipeline of One-step Personalization in \ourmethod}
\label{alg:training}
\begin{algorithmic}[1]
\State \textbf{Input:} Real image dataset $\mathcal{D} = \{(x_0^r, y)\}$; teacher model $\model^{tc}$; student model $\sdmodel^{st}$; discriminators $\{\mathcal{D}_k\}_{k=1}^{K}$; text encoder $\conditioner(\cdot)$; diffusion steps $T$; noise schedule functions $\{\alpha_t, \sigma_t\}$; weighting functions $c(t)$, $\lambda_{\text{id}}, \lambda_{\text{mse}}, \lambda_{\text{ms}}, \{\lambda_k\}_{k=1}^K$

\For{each training iteration}
    \State Sample real image and prompt: $(x_0^r, y) \sim \mathcal{D}$
    \State Random noise $\epsilon \sim \mathcal{N}(0,1)$
    \State Encode prompt: $\textembedding \gets \conditioner(y)$
    \State \textbf{Step 1: Teacher training}
    \State Sample timestep $t \sim \mathcal{U}(1, T)$
    \State Generate noisy input: $x_t = \alpha_t x_0^r + \sigma_t \epsilon$
    \State Predict noise: $\hat{\epsilon} \gets \model^{tc}(x_t, t, \textembedding)$
    \State Compute loss: $\mathcal{L}_{\text{rec}} = \| \epsilon - \hat{\epsilon} \|_2^2$
    \State Update teacher model and the text encoder using $\mathcal{L}_{\text{rec}}$
    
    \State \textbf{Step 2: Student training}
    \State Sample latent: $x_T \sim \mathcal{N}(0,1)$
    \State Generate image: $x_0^{\text{st}} \gets \sdmodel^{st}(x_T, T, \text{stopgrad}(\textembedding))$ \Comment{$\text{stopgrad}(\cdot)$ denotes stop-gradient}

    \State \textit{// Alignment loss}
    \State Forward diffuse: $x_t^{\text{st}} = \alpha_t x_0^{\text{st}} + \sigma_t \epsilon$
    \State Denoise: $x_0^{\text{tc}} \gets \text{stopgrad}(\model^{tc}(x_t^{\text{st}}, t, \textembedding))$
    \State Compute alignment loss: 
    \Statex \hspace{1.5em}
    $\mathcal{L}_{\text{align}} = c(t) \cdot \left[ 
    \lambda_{\text{id}} \cdot \mathcal{L}_{\text{id}}(x_0^{st}, x_0^{tc}) +
    \lambda_{\text{mse}} \cdot \mathcal{L}_{\text{mse}}(x^s, x^t) +
    \lambda_{\text{ms}} \cdot \mathcal{L}_{\text{swd}}(x_0^{st}, x_0^{tc})
    \right]$

    \State \textit{// Adversarial loss}
    \State Compute adversarial loss: $\mathcal{L}_{\text{GAN}}^{G} = \sum_{k=1}^{K} \lambda_k \cdot \mathbb{E}_{x_0^{st}} \left[ -\log(D_k(x_0^{st})) \right]$
    
    \State Update student model using: $\mathcal{L}_{\text{st}} = \mathcal{L}_{\text{align}} + \mathcal{L}_{\text{GAN}}^{G}$
    
    \State \textbf{Step 3: Discriminator training}
    \For{each discriminator $\mathcal{D}_k$}
        \State Compute the discriminator loss: 
        \Statex \hspace{3em}
        $\mathcal{L}_{\text{GAN}}^{D_k} = -\left[ \mathbb{E}_{x_0^r} \left[ \log D_k(x_0^r) \right] + \mathbb{E}_{x_0^{st}} \left[ \log(1 - D_k(\text{stopgrad}(x_0^{st}))) \right] \right]$
        \State Update $\mathcal{D}_k$ using: $\mathcal{L}_{\text{GAN}}^{D_k}$
    \EndFor
\EndFor
\State \textbf{Output:} Trained teacher model $\model^{tc}$, student model $\sdmodel^{st}$, and text encoder $\conditioner$
\end{algorithmic}
\end{algorithm*}

\minisection{Collaborative Learning Stage.}
During the collaborative learning stage, the training remains the same as before, except that the training examples are replaced with one-step inference samples generated by the student model. This design is beneficial in multiple ways. \textit{(1)} In the initial distillation stage, the availability of real images is limited, and this constrained data scale impedes the training of the teacher model. By contrast, the student model exhibits the ability to learn data distributions from few images. That is a capability endowed by the discriminator, as validated in few-shot GAN frameworks such as TransferGAN\citep{Wang2018TransferringGG} and MineGAN\citep{wang2020minegan}. Consequently, during the collaborative learning stage, our objective is to sample from the data distribution learned by the student model, using these samples as training examples to enhance the teacher model’s performance. 
\textit{(2)} Notably, the 1-step diffusion student model learns distributions distinct from those acquired by the teacher model. Similar observation can be found in ADD\citep{sauer2023adversarial}, where the discriminator loss primarily shapes the data distribution of the student model, while the distillation loss facilitates convergence and enhances conceptual alignment with the teacher’s outputs. Supporting evidence for this ablation study can be found in Table~\ref{tab:ablation_study_full}. 
\textit{(3)} By shifting the training examples from few real image inputs to images generated by the 1-step student model, the teacher model is enabled to learn from the student's distribution, a distribution partially shaped by the discriminator’s design and characterized by image features not inherently present in the teacher model, thereby yielding beneficial effects.

\section{Additional results on method comparison}
\label{appendix:comparison}

\begin{table*}[t]
\caption{Quantitative comparisons with existing text-to-image (T2I) personalization methods. NFEs indicates the number of function evaluations.}
\resizebox{1.0\linewidth}{!}{
\begin{tabular}{c|c|cc|ccc|cc|c}
\toprule
\multirow{2}{*}{\textbf{\tabincell{c}{Methods}}}  & \multirow{2}{*}{\textbf{\tabincell{c}{Model}}}  & \multirow{2}{*}{\textbf{\tabincell{c}{Train \\ NFEs}}}  & \multirow{2}{*}{\textbf{\tabincell{c}{Inference \\ NFEs}}}   & \multirow{2}{*}{\textbf{\tabincell{c}{CLIP-T$\uparrow$}}}   & \multirow{2}{*}{\textbf{\tabincell{c}{CLIP-I$\uparrow$}}}    & \multirow{2}{*}{\textbf{\tabincell{c}{DINO$\uparrow$}}}    & \multirow{2}{*}{\textbf{\tabincell{c}{Train \\ Time (s)}}} & \multirow{2}{*}{\textbf{\tabincell{c}{Inference \\ Time (s)}}} & \multirow{2}{*}{\textbf{\tabincell{c}{iterations}}}  \\
&  & & & & & & & \\ 

\midrule
\multirow{7}{*}{Custom diffusion}  & SD 2.1  & 1000  & 25  & 0.264  & 0.761  & 0.555  &  345  & 2.73 & 1000 \\ 
                 & SD Turbo & 1000    & 1    & 0.207  & 0.530  & 0.097  & 541  & 0.22 & 1000 \\
                 & SD Turbo & 1000    & 4    & 0.257  & 0.597  & 0.235  & 541  & 0.54 & 1000 \\
                 & SD Turbo & 1000    & 25   & 0.276  & 0.647  & 0.337  & 541  & 1.57 & 1000 \\
                 
                 & SD Turbo & 1       & 1    & 0.205  & 0.518  & 0.058  & 543  & 0.22 & 1000 \\
                 & SD Turbo & 1       & 4    & 0.246  & 0.556  & 0.109  & 543  & 0.53 & 1000 \\
                 & SD Turbo & 4       & 1    & 0.206  & 0.532  & 0.105  & 543  & 0.23 & 1000 \\
                 & SD Turbo & 4       & 4    & 0.258  & 0.600  & 0.244  & 543  & 0.54 & 1000 \\
\midrule
Textual Inversion  & SD Turbo & 1 & 1 & 0.252  & 0.564  & 0.166  & 2269 & 0.13 & 4000 \\ 
Cones 2            & SD Turbo & 1 & 1 & 0.273  & 0.619  & 0.204  & 2446 & 0.51 & 4000 \\
DreamBooth         & SD Turbo & 1 & 1 & 0.188  & 0.536  & 0.111  & 281  & 0.14 & 1000 \\
TextBoost          & SD Turbo & 1 & 1 & 0.217  & 0.570  & 0.167  & 64   & 0.15 & 500 \\
DisenBooth         & SD Turbo & 1 & 1 & 0.251  & 0.564  & 0.231  & 905  & 0.18 & 2000 \\
Lora               & SD Turbo & 1 & 1 & 0.212  & 0.585  & 0.160  & 141  & 0.15 & 800 \\
IP-Adapter         & TCD + SDXL     & / & 1 & 0.204  & 0.628  & 0.325  &  /   & 0.39 & /  \\
OminiControl       & Flux     & / & 1 & \textbf{0.279} & 0.727 & 0.455  & / & 2.48 & / \\

\midrule 

\ourmethod & SD Turbo & 1 & 1 & 0.252 & \textbf{0.783} & \textbf{0.637} & 3137 & 0.18 & 1000 \\
\bottomrule    

\end{tabular}
}
\label{tab:main_quantitative_results_appendix}
\vspace{-4mm}
\end{table*}

\begin{figure*}[t]
    \centering
    \includegraphics[width=0.8\linewidth]{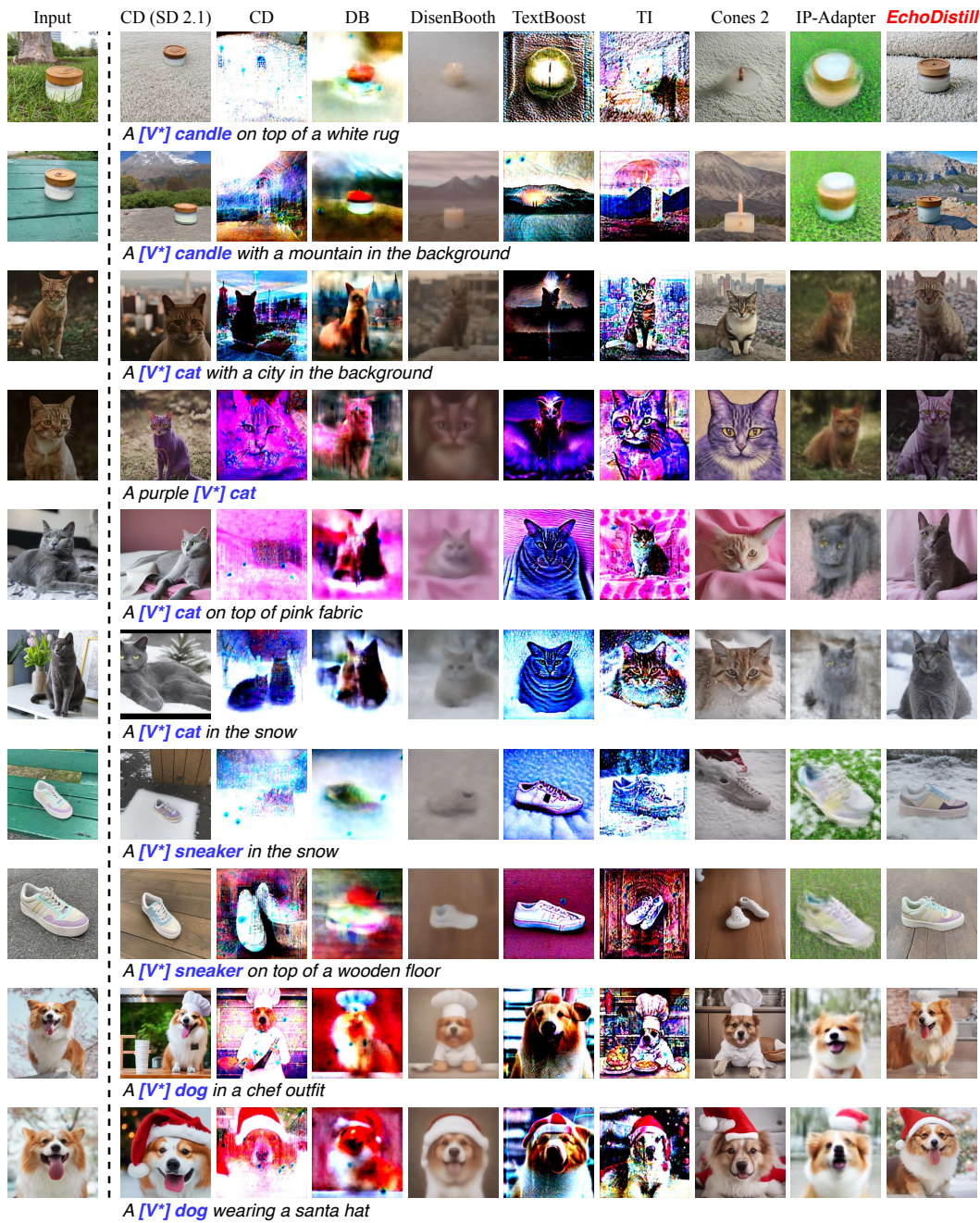}
    \caption{Our method \ourmethod (last column) compared with existing methods applied to the \ourproblem setup with SDTurbo~\citep{sauer2023adversarial} as the one-step diffusion backbone. One representive concept image is shown on the left-most column. (Part 1)
    }
    \label{fig:comparisons_supp1}
\end{figure*}

\begin{figure*}[t]
    \centering
    \includegraphics[width=0.8\linewidth]{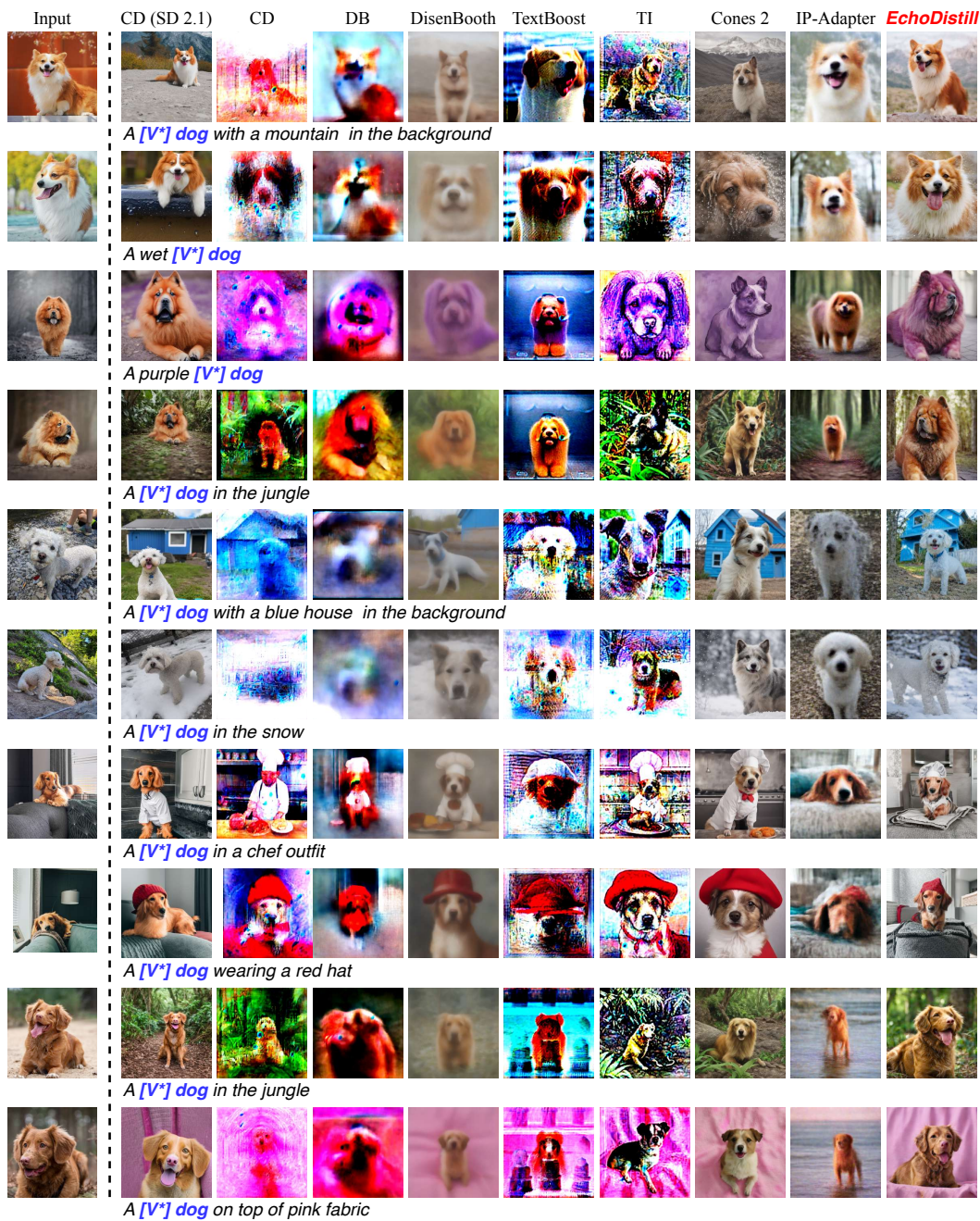}
    \caption{Our method \ourmethod (last column) compared with existing methods applied to the \ourproblem setup with SDTurbo~\citep{sauer2023adversarial} as the one-step diffusion backbone. One representive concept image is shown on the left-most column. (Part 2)
    }
    \label{fig:comparisons_supp2}
\end{figure*}

In this section, we present additional quantitative and qualitative results to validate the effectiveness and efficiency of our proposed method.
Table~\ref{tab:main_quantitative_results_appendix} extends the comparisons from the main paper by reporting both training and inference time (in seconds), along with the number of optimization iterations required by each method. These results underscore the efficiency of our approach, achieving image generation in just \textit{0.18 seconds} per instance during inference.
Figures~\ref{fig:comparisons_supp1} through~\ref{fig:comparisons_supp4} provide additional qualitative comparisons against representative baseline methods. Each figure presents a concept reference image (left) followed by results from various approaches. Our method, \ourmethod, consistently delivers superior visual fidelity while maintaining rapid inference, highlighting its practical advantages for resource-constrained applications in one-step diffusion-based image generation (\ourproblem).

\minisection{Discussion on runtime cost.} About the training time, this limitation is inherent to optimization-based customization approaches, which universally require additional runtime computation when encountering novel concepts. Conversely, existing optimization-free methods, including encoder-based frameworks (IP-Adatper~\citep{ye2023ip-adapter}, DreamO~\citep{mou2025dreamo}, Xverse~\citep{chen2025xverse}, UNO~\citep{wu2025less}, InfiniteYou~\citep{jiang2025infiniteyou}, etc.) and unified models (BAGEL~\citep{deng2025bagel}, GPT-4o~\citep{achiam2023gpt}, OmniGen2~\citep{wu2025omnigen2}, etc.), demand extensive datasets for training. Furthermore, no encoder-based or unified model to date fully supports few-step (or even one-step) diffusion models. This leaves integrating one-step speed with unified model versatility largely underexplored. In this work, we aim to be the first to investigate the realization of one-step personalization via an optimization-based approach, with optimization-free alternatives designated as future work.

\minisection{Comparison with Flux+OminiControl.}We further compare \ourmethod with the recent Flux~\citep{flux2024} model combined with the OminiControl~\citep{tan2025ominicontrol}. It is important to note that OminiControl is trained on large-scale datasets similar to IP-Adapter, which makes the evaluation against our method not entirely equitable. The evaluation is conducted on the DreamBooth dataset under the 1-step setup, and the results are summarized in the lower part of Table~\ref{tab:main_quantitative_results_appendix}. As shown, Flux+OminiControl outperforms other baselines reported in the table; however, it remains significantly inferior to our proposed \ourmethod. This performance gap can be attributed to the weaker generation capability of Flux constrained to 1-step inference.

\section{Discussions for alternative designs}
\label{appendix:Additional discussions for alternative designs}
We further compare our method with several alternative designs in order to clarify the motivation and validity of our proposed framework. All experiments in this section are performed on the DreamBooth dataset.

\textbf{Teacher-first paradigm.}
In this design, the teacher is first trained, and the teacher-generated samples for the target concept (with varying text prompts) are directly used as supervised training data for the student. The identity loss (Eq.~\ref{eq:id_loss} in the main paper) is applied between the teacher-generated samples and the student outputs, allowing the student to learn identity-related features from the teacher model. As shown in Table~\ref{tab:alt_designs}, this design performs worse than our proposed approach. Moreover, it suffers from several inherent limitations: (1) Computational overhead: teacher inference requires multiple steps, which is inefficient; (2) Teacher irreliability: as discussed in the main paper, the teacher does not always successfully learn the target concepts; (3) Limited image diversity: the generated images consistently feature highly similar visual appearances;  and (4) Performance ceiling: the student’s performance is inherently bounded by the capabilities of the teacher.

In addition, we also attempted to directly apply VSD ~\citep{wang2023prolificdreamer}, SDS~\citep{poole2023dreamfusion}, and MSE losses to distill teacher-learned concepts into the student model under this paradigm. However, we observed that this approach was insufficient for transferring the teacher’s personalization capabilities to the student.

\textbf{Discussion on feed-forward customization methods.}
Beyond the teacher-first paradigm, an alternative direction is to build on feed-forward customization methods such as SynCD~\citep{kumari2025generating} and JeDi~\citep{zeng2024jedi}, and then distill these models into a few-step diffusion framework. However, most existing distillation techniques for diffusion models are primarily designed to align the data distributions of few-step models with those of their teacher models. This emphasis stems from the inherent difficulty that few-step diffusion models face in replicating the full denoising trajectory of their teacher. As a result, subtle discrepancies in concept-specific details are often introduced, as observed in prior works such as AYF~\citep{sabour2025align}, ADD~\citep{sauer2023adversarial}, and LCM~\citep{luo2023latent}.

\textbf{Remove STE.}
We further investigate the effect of sharing the text encoder between the teacher and student models. Removing the shared text encoder (STE) results in a clear performance drop, demonstrating that STE not only simplifies the training framework but also improves learning efficiency.

\begin{table}[t]
\centering
\caption{Additional results for alternative designs on the DreamBooth dataset.}
\label{tab:alt_designs}
\begin{tabular}{lccc}
\toprule
Method & CLIP-T $\uparrow$ & CLIP-I $\uparrow$ & DINO $\uparrow$ \\
\midrule
Teacher-first & \textbf{0.266} & 0.725 & 0.503 \\
Remove STE & 0.242 & 0.742 & 0.551 \\
Ours (\ourmethod) & 0.252 & \textbf{0.783} & \textbf{0.637} \\
\bottomrule
\end{tabular}
\end{table}

\section{User study}

\begin{table}[t]
\centering
\caption{User study for diverse methods.}
\label{tab: User study}
\begin{tabular}{cc}
\toprule
Method           & Preference Rate (\%)    \\
\midrule 
Custom Diffusion \citep{kumari2023customdiffusion} & 32.26\% \\
Cones 2 \citep{Cones2023}           & 0.48\%  \\
DisenBooth \citep{chen2023disenbooth}       & 0.36\%  \\
\ourmethod             & 66.90\% \\
\bottomrule
\end{tabular}
\end{table}

\begin{table}[t]
\centering
\caption{One-shot performance of our \ourmethod.}
\begin{tabular}{c|ccc}
\toprule
\textbf{Methods} & \textbf{CLIP-T} & \textbf{CLIP-I} & \textbf{DINO} \\
\midrule
One-shot  & 0.231           & 0.713          & 0.470          \\
Few-shot  & \textbf{0.252} & \textbf{0.783} & \textbf{0.637} \\
\bottomrule
\end{tabular}
\label{tab:One-shot}
\end{table}

To evaluate alignment with human preferences, we conducted a user study involving 15 participants, yielding 840 preference annotations per method. In each trial, participants were presented with a set of generated images and instructed to “select the best image from each group, considering both text-image alignment and object identity consistency.” The methods evaluated in our study include Custom Diffusion (under a multi-step setting), as well as Cones 2, DisenBooth, and our method (all under a single-step setting).
As summarized in Table~\ref{tab: User study}, our approach, \ourmethod, significantly outperformed other methods, receiving at least 34\% more user votes than the second-best method. These findings underscore the strong alignment between \ourmethod’s outputs and human perceptual judgments.

\section{Extended Ablation Study}
\begin{figure*}[t]
    \centering
    \includegraphics[width=0.8\linewidth]{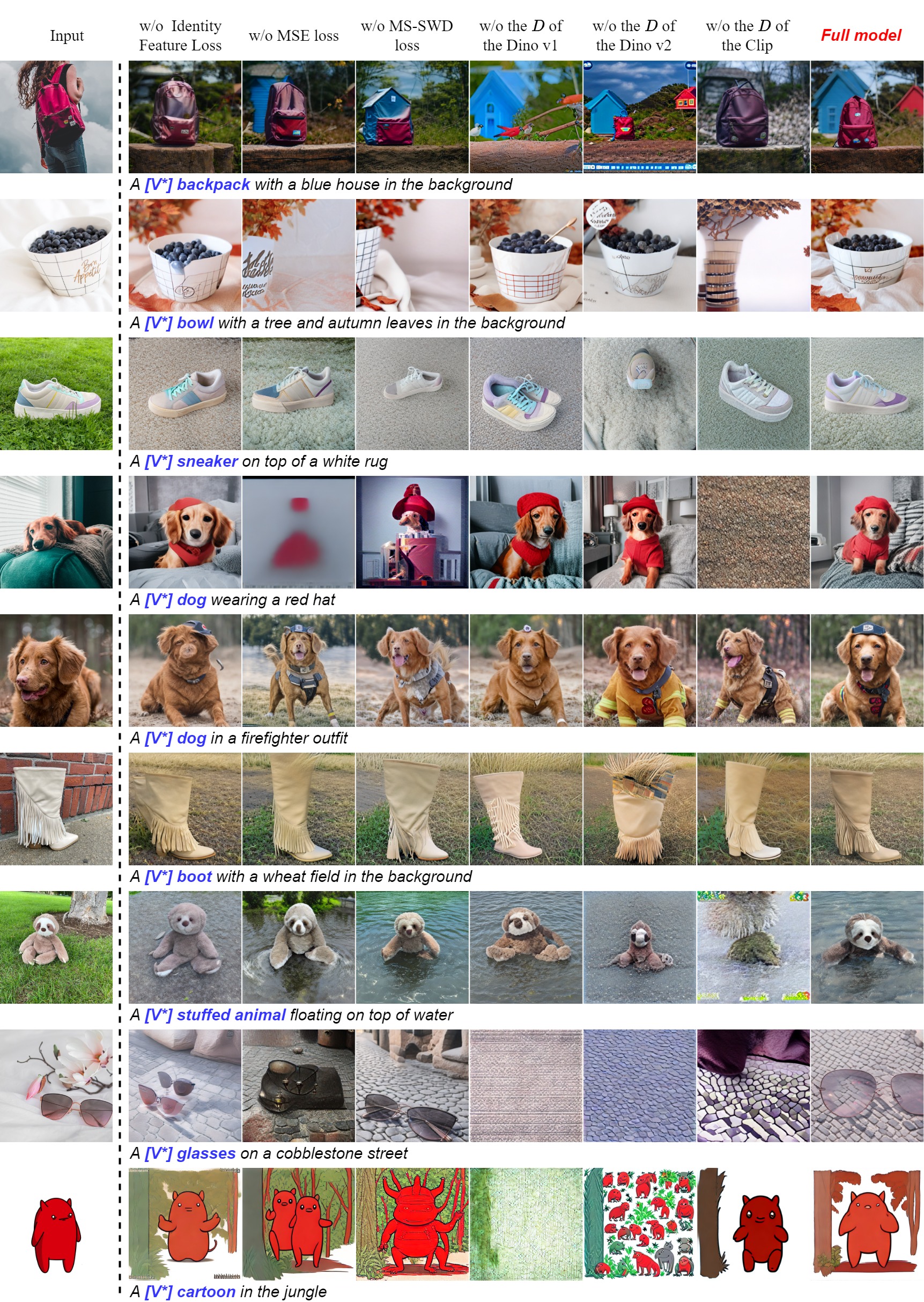}
    \caption{Qualitative results of the extended ablation study. $D$ denotes the discriminator.
    }
    \label{fig:ablation_qual_suppl}
\end{figure*}

In Sec.~\ref{Ablation Study} of the main paper, we explore the contributions of key components in \ourmethod, namely the teacher model and discriminators. 
A more detailed ablation study is presented in Table~\ref{tab:ablation_study_full}, wherein individual loss terms and discriminators are systematically removed. The results indicate that omitting the ID loss, latent MSE loss, or MSSWD loss causes notable performance degradation, particularly reflected in reduced DINO scores, underscoring their critical role in maintaining alignment with the teacher model. 
Furthermore, removal of any single discriminator leads to more pronounced declines across all evaluation metrics. Collectively, these findings demonstrate the complementary nature of the various loss functions and discriminators in improving generation fidelity and semantic consistency. 
Qualitative comparisons provided in Fig.~\ref{fig:ablation_qual_suppl} further illustrate the visual impact of removing each component. Beyond the general decline in generation quality, we observe that omission of certain components can induce training instability or divergence for specific concepts.

\begin{figure}[t!]
    \centering
    \includegraphics[width=0.99\linewidth]{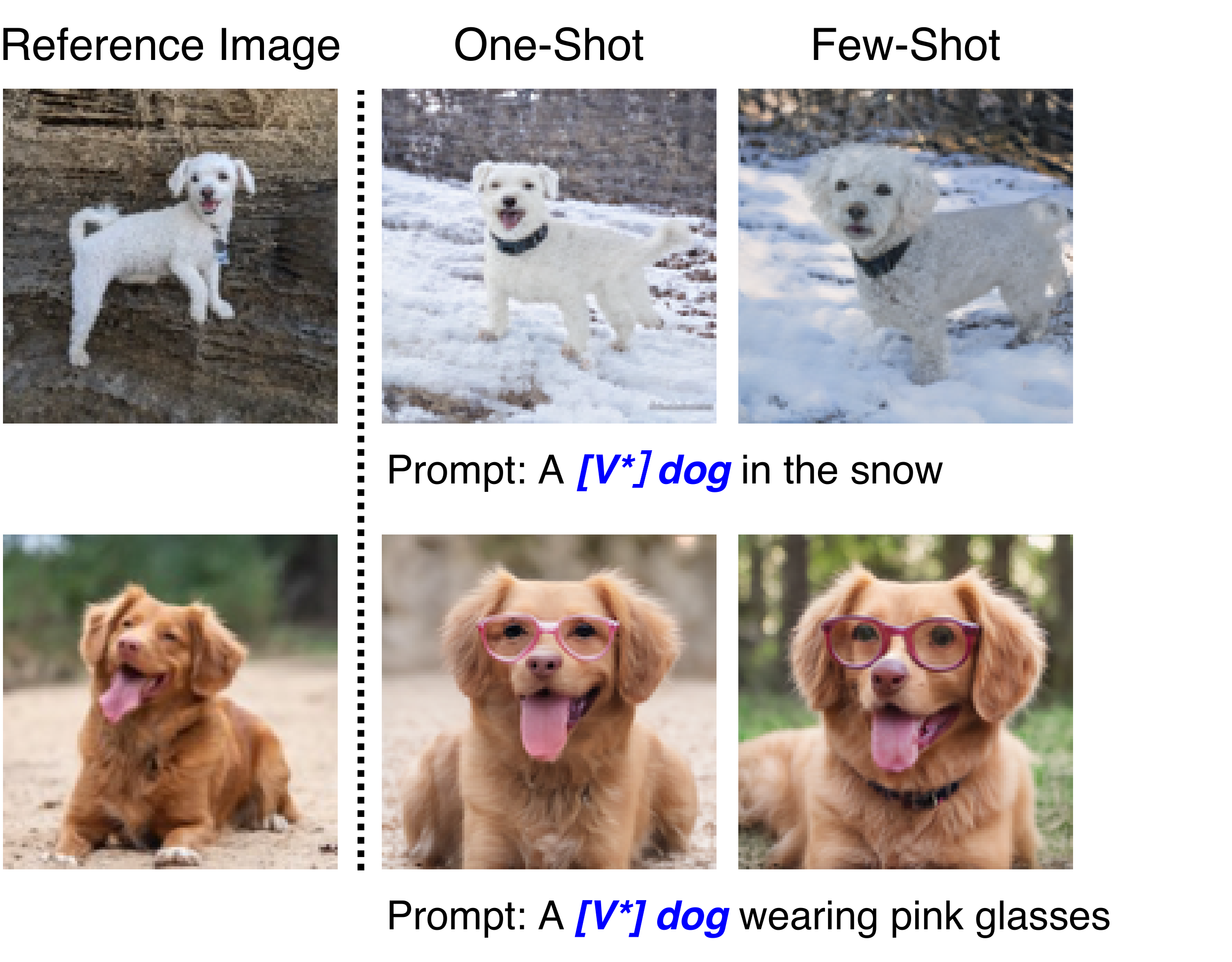}
    \vspace{-3mm}
    \caption{The qualitative results of the 1-shot performance.}
    \label{fig:one_shot}
\end{figure}

\begin{table}[t]
\centering
\caption{Ablation study of our method \ourmethod.}
\resizebox{1.01\linewidth}{!}{
\begin{tabular}{c|cccc}
\toprule
\textbf{Methods}       &   \textbf{CLIP-T} &  \textbf{CLIP-I} &  \textbf{DINO}   \\
\midrule
Full model             & \textbf{0.252}     & \textbf{0.783}   & \textbf{0.637} \\
\midrule
w/o the teacher  & 0.240 &  0.719 &   0.505 \\
w/o all the discriminators &  0.200 & 0.566 & 0.105 \\
\midrule
w/o  Identity Feature Loss    & 0.248        & 0.769   & 0.618 \\
w/o MSE loss       & 0.242        & 0.739   & 0.528 \\
w/o MS-SWD loss            & 0.249        & 0.754   & 0.553 \\
w/o the discriminator of the Dino v1 & 0.231        & 0.689   & 0.441 \\
w/o the discriminator of the Dino v2 & 0.246        & 0.736   & 0.534 \\
w/o the discriminator of the Clip    & 0.227        & 0.678   & 0.409 \\    
\bottomrule    
\end{tabular}
}
\label{tab:ablation_study_full}
\end{table}

\section{1-shot performance.}
We further evaluate our method under a one-shot supervision setting, wherein only a single image is utilized for training. As summarized in Table~\ref{tab:One-shot}, performance declines across all evaluation metrics relative to the few-shot scenario. This degradation is anticipated, given that our approach is not explicitly optimized for one-shot learning, and the scarcity of supervisory data increases the likelihood of training instability. Qualitative results illustrated in Fig.~\ref{fig:one_shot} demonstrate that, although one-shot training can yield visually plausible outputs, the generated images occasionally lack fine-grained details corresponding to the novel concept.

\section{Results on the CustomConcept101 Dataset}
\label{CustomConcept101 Dataset}
\begin{figure*}[t]
    \centering
    \includegraphics[width=0.8\linewidth]{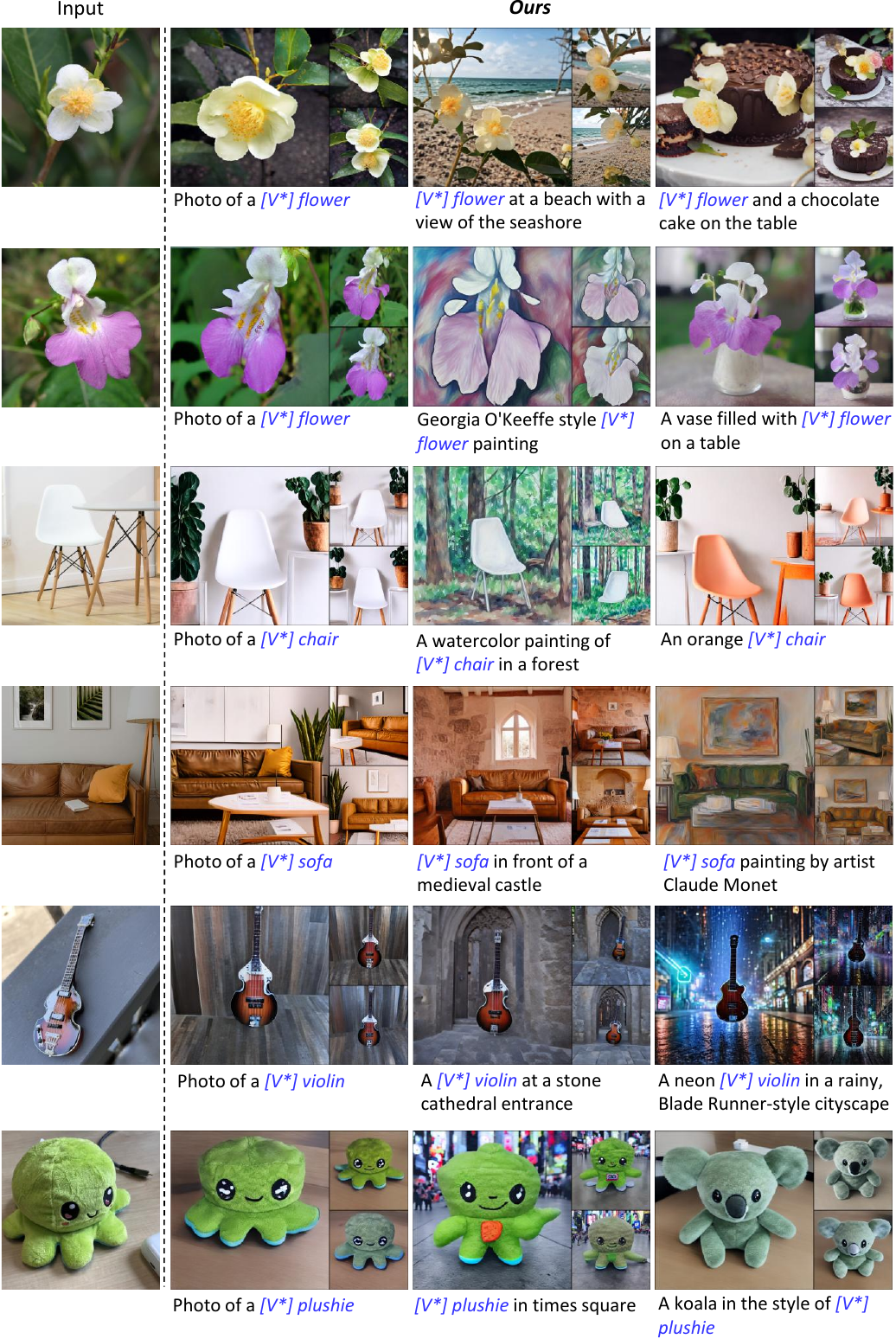}
    \caption{Qualitative results of \ourmethod on the CustomConcept101 dataset. Our method demonstrates strong generalization across a variety of concepts and prompt styles. (Part 1)
    }
    \label{fig:CustomConcept101_1}
\end{figure*}

\begin{figure*}[t]
    \centering
    \includegraphics[width=0.8\linewidth]{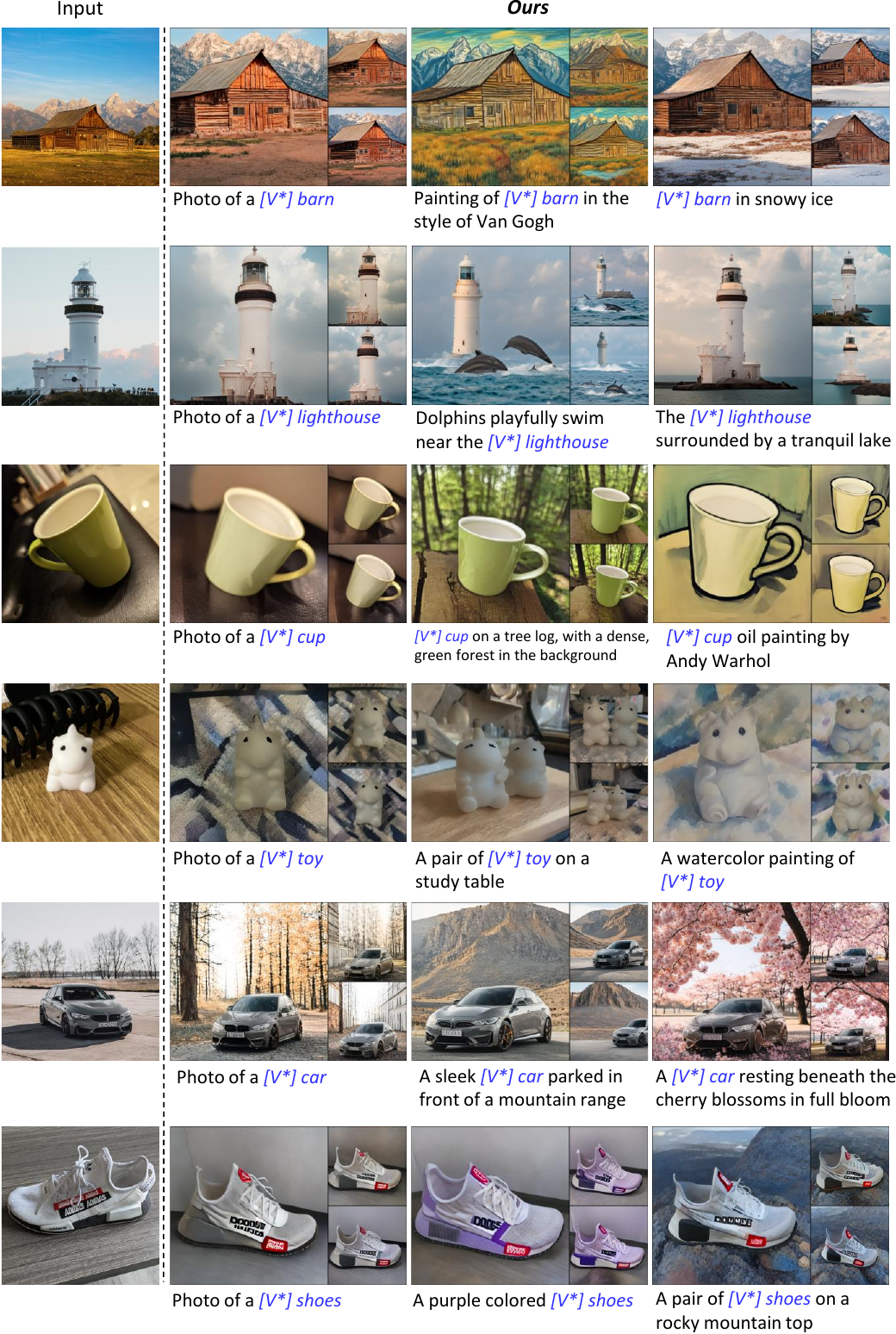}
    \caption{Qualitative results of \ourmethod on the CustomConcept101 dataset. Our method demonstrates strong generalization across a variety of concepts and prompt styles. (Part 2). 
    }
    \label{fig:CustomConcept101_2}
\end{figure*}

We further evaluate our method on the CustomConcept101 dataset~\citep{kumari2023customdiffusion}. Qualitative results, presented in Fig.~\ref{fig:CustomConcept101_1} and Fig.~\ref{fig:CustomConcept101_2}, demonstrate that our approach generalizes effectively across a diverse set of concepts and prompt types, consistently generating high-quality outputs.

\clearpage

\begin{figure*}[t]
    \centering
    \includegraphics[width=0.8\linewidth]{figs/supp/supp3_compressed.pdf}
    \caption{Our method \ourmethod (last column) compared with existing methods applied to the \ourproblem setup with SDTurbo~\citep{sauer2023adversarial} as the one-step diffusion backbone. One representive concept image is shown on the left-most column. (Part 3)
    }
    \label{fig:comparisons_supp3}
\end{figure*}

\begin{figure*}[t]
    \centering
    \includegraphics[width=0.8\linewidth]{figs/supp/supp4_compressed.pdf}
    \caption{Our method \ourmethod (last column) compared with existing methods applied to the \ourproblem setup with SDTurbo~\citep{sauer2023adversarial} as the one-step diffusion backbone. One representive concept image is shown on the left-most column. (Part 4)
    }
    \label{fig:comparisons_supp4}
\end{figure*}

%% file: longstrings.bib
@string{ai   =  "Artificial Intelligence"}

@string{aaai = "Proceedings of the Conference on Artificial Intelligence"}

@string{cvpr   = "Proceedings of the IEEE Conference on Computer Vision and Pattern Recognition"}

@string{eccv  = "Proceedings of the European Conference on Computer Vision"}

@string{eccv  = "European Conference on Computer Vision"}

@string{iccv =  "Proceedings of the International Conference on Computer Vision"}

@string{icml =   "International Conference on Machine Learning"}

@string{ijcv =  "International Journal of Computer Vision"}

@string{nips = "Advances in Neural Information Processing Systems"}

@string{neurips = "Advances in Neural Information Processing Systems"}

@string{siggraph = "Proceedings of the ACM SIGGRAPH Conference on Computer Graphics"}

@string{wacv =  "Proceedings of the {IEEE} Workshop on Applications of Computer Vision"}

@string{ICLR = "International Conference on Learning Representations"}

@string{jun = "June"}


%% file: main.bib
@article{tian2024VAR,
  title={Visual autoregressive modeling: Scalable image generation via next-scale prediction},
  author={Tian, Keyu and Jiang, Yi and Yuan, Zehuan and Peng, Bingyue and Wang, Liwei},
  journal={Advances in neural information processing systems},
  volume={37},
  pages={84839--84865},
  year={2024}
}

@article{han2024infinity,
  title={Infinity: Scaling bitwise autoregressive modeling for high-resolution image synthesis},
  author={Han, Jian and Liu, Jinlai and Jiang, Yi and Yan, Bin and Zhang, Yuqi and Yuan, Zehuan and Peng, Bingyue and Liu, Xiaobing},
  journal=cvpr,
  year={2025}
}

@article{wu2025proxy_tuning_ar,
  title={Proxy-Tuning: Tailoring Multimodal Autoregressive Models for Subject-Driven Image Generation},
  author={Wu, Yi and Zhu, Lingting and Liu, Lei and Qiao, Wandi and Li, Ziqiang and Yu, Lequan and Li, Bin},
  journal={arXiv preprint arXiv:2503.10125},
  year={2025}
}

@article{chung2025finetune_var,
  title={Fine-Tuning Visual Autoregressive Models for Subject-Driven Generation},
  author={Jiwoo Chung and Sangeek Hyun and Hyunjun Kim and Eunseo Koh and MinKyu Lee and Jae-Pil Heo},
  journal=iccv,
  year={2025}
}

@article{garibi2025tokenverse,
  title={TokenVerse: Versatile Multi-concept Personalization in Token Modulation Space},
  author={Garibi, Daniel and Yadin, Shahar and Paiss, Roni and Tov, Omer and Zada, Shiran and Ephrat, Ariel and Michaeli, Tomer and Mosseri, Inbar and Dekel, Tali},
  journal=iccv,
  year={2025}
}

@article{xu2025show_o_turbo,
  title={Show-o Turbo: Towards Accelerated Unified Multimodal Understanding and Generation},
  author={Xu, Chenkai and Wang, Xu and Liao, Zhenyi and Li, Yishun and Hou, Tianqi and Deng, Zhijie},
  journal={arXiv preprint arXiv:2502.05415},
  year={2025}
}

@article{liu2024distilled_DD,
  title={Distilled decoding 1: One-step sampling of image auto-regressive models with flow matching},
  author={Liu, Enshu and Ning, Xuefei and Wang, Yu and Lin, Zinan},
  journal=iclr,
  year={2025}
}

@article{sun2024llamagen,
  title={Autoregressive model beats diffusion: Llama for scalable image generation},
  author={Sun, Peize and Jiang, Yi and Chen, Shoufa and Zhang, Shilong and Peng, Bingyue and Luo, Ping and Yuan, Zehuan},
  journal={arXiv preprint arXiv:2406.06525},
  year={2024}
}

@article{wang2024instantid,
  title={Instantid: Zero-shot identity-preserving generation in seconds},
  author={Wang, Qixun and Bai, Xu and Wang, Haofan and Qin, Zekui and Chen, Anthony},
  journal={arXiv preprint arXiv:2401.07519},
  year={2024}
}

@article{li2025comprehensive_survey,
  title={A Comprehensive Survey on Visual Concept Mining in Text-to-image Diffusion Models},
  author={Li, Ziqiang and Li, Jun and Xiong, Lizhi and Fu, Zhangjie and Li, Zechao},
  journal={arXiv preprint arXiv:2503.13576},
  year={2025}
}

@article{xiao2023fastcomposer,
  title={Fastcomposer: Tuning-free multi-subject image generation with localized attention},
  author={Xiao, Guangxuan and Yin, Tianwei and Freeman, William T and Durand, Fr{\'e}do and Han, Song},
  journal=ijcv,
  year={2023}
}

@article{ye2023ip-adapter,
  title={Ip-adapter: Text compatible image prompt adapter for text-to-image diffusion models},
  author={Ye, Hu and Zhang, Jun and Liu, Sibo and Han, Xiao and Yang, Wei},
  journal=aaai,
  year={2024}
}

@inproceedings{Wang2018TransferringGG,
  title={Transferring GANs: generating images from limited data},
  author={Yaxing Wang and Chenshen Wu and Luis Herranz and Joost van de Weijer and Abel Gonzalez-Garcia and B. Raducanu},
  booktitle={ECCV},
  year={2018}
}

@article{huang2024classdiffusion,
  title={Classdiffusion: More aligned personalization tuning with explicit class guidance},
  author={Huang, Jiannan and Liew, Jun Hao and Yan, Hanshu and Yin, Yuyang and Zhao, Yao and Shi, Humphrey and Wei, Yunchao},
  journal=iclr,
  year={2025}
}

@article{pang2024attndreambooth,
  title={Attndreambooth: Towards text-aligned personalized text-to-image generation},
  author={Pang, Lianyu and Yin, Jian and Zhao, Baoquan and Wu, Feize and Wang, Fu Lee and Li, Qing and Mao, Xudong},
  journal={Advances in Neural Information Processing Systems},
  volume={37},
  pages={39869--39900},
  year={2024}
}

@article{zhou2023enhancing_profusion,
  title={Enhancing detail preservation for customized text-to-image generation: A regularization-free approach},
  author={Zhou, Yufan and Zhang, Ruiyi and Sun, Tong and Xu, Jinhui},
  journal={arXiv preprint arXiv:2305.13579},
  year={2023}
}

@article{huang2024resolving_rmcc,
  title={Resolving multi-condition confusion for finetuning-free personalized image generation},
  author={Huang, Qihan and Fu, Siming and Liu, Jinlong and Jiang, Hao and Yu, Yipeng and Song, Jie},
  journal=aaai,
  year={2025}
}

@inproceedings{ding2024freecustom,
  title={Freecustom: Tuning-free customized image generation for multi-concept composition},
  author={Ding, Ganggui and Zhao, Canyu and Wang, Wen and Yang, Zhen and Liu, Zide and Chen, Hao and Shen, Chunhua},
  booktitle={Proceedings of the IEEE/CVF Conference on Computer Vision and Pattern Recognition},
  pages={9089--9098},
  year={2024}
}

@article{park2024textboost,
  title={TextBoost: Towards One-Shot Personalization of Text-to-Image Models via Fine-tuning Text Encoder},
  author={Park, NaHyeon and Kim, Kunhee and Shim, Hyunjung},
  journal={arXiv preprint arXiv:2409.08248},
  year={2024}
}

@article{wang2024ms_diffusion,
  title={Ms-diffusion: Multi-subject zero-shot image personalization with layout guidance},
  author={Wang, Xierui and Fu, Siming and Huang, Qihan and He, Wanggui and Jiang, Hao},
  journal=iclr,
  year={2025}
}

@inproceedings{kong2024omg,
  title={Omg: Occlusion-friendly personalized multi-concept generation in diffusion models},
  author={Kong, Zhe and Zhang, Yong and Yang, Tianyu and Wang, Tao and Zhang, Kaihao and Wu, Bizhu and Chen, Guanying and Liu, Wei and Luo, Wenhan},
  booktitle={European Conference on Computer Vision},
  pages={253--270},
  year={2024},
  organization={Springer}
}

@inproceedings{po2024orthogonal,
  title={Orthogonal adaptation for modular customization of diffusion models},
  author={Po, Ryan and Yang, Guandao and Aberman, Kfir and Wetzstein, Gordon},
  booktitle={Proceedings of the IEEE/CVF Conference on Computer Vision and Pattern Recognition},
  pages={7964--7973},
  year={2024}
}

@misc{flux2024,
    author={Black Forest Labs},
    title={FLUX},
    year={2024},
    howpublished={\url{https://github.com/black-forest-labs/flux}},
}

@inproceedings{tumanyan2023plug,
  title={Plug-and-play diffusion features for text-driven image-to-image translation},
  author={Tumanyan, Narek and Geyer, Michal and Bagon, Shai and Dekel, Tali},
  booktitle={Proceedings of the IEEE/CVF Conference on Computer Vision and Pattern Recognition},
  pages={1921--1930},
  year={2023}
}

@misc{rowles2024ipadapterinstruct,
      title={IPAdapter-Instruct: Resolving Ambiguity in Image-based Conditioning using Instruct Prompts}, 
      author={Ciara Rowles and Shimon Vainer and Dante De Nigris and Slava Elizarov and Konstantin Kutsy and Simon Donné},
      year={2024},
      eprint={2408.03209},
      archivePrefix={arXiv},
      primaryClass={cs.CV},
      url={https://arxiv.org/abs/2408.03209}, 
}

@InProceedings{Rombach_2022_CVPR_stablediffusion,
    author    = {Rombach, Robin and Blattmann, Andreas and Lorenz, Dominik and Esser, Patrick and Ommer, Bj\"orn},
    title     = {High-Resolution Image Synthesis With Latent Diffusion Models},
    booktitle = {Proceedings of the IEEE/CVF Conference on Computer Vision and Pattern Recognition (CVPR)},
    month     = {06},
    year      = {2022},
    pages     = {10684-10695}
}

@article{saharia2022imagen,
  title={Photorealistic text-to-image diffusion models with deep language understanding},
  author={Saharia, Chitwan and Chan, William and Saxena, Saurabh and Li, Lala and Whang, Jay and Denton, Emily and Ghasemipour, Seyed Kamyar Seyed and Ayan, Burcu Karagol and Mahdavi, S Sara and Lopes, Rapha Gontijo and others},
  journal=nips,
  year={2022}
}

@article{hertz2022prompt,
  title={Prompt-to-prompt image editing with cross attention control},
  author={Hertz, Amir and Mokady, Ron and Tenenbaum, Jay and Aberman, Kfir and Pritch, Yael and Cohen-Or, Daniel},
  journal=iclr,
  year={2023}
}

@inproceedings{radford2021clip,
  title={Learning transferable visual models from natural language supervision},
  author={Radford, Alec and Kim, Jong Wook and Hallacy, Chris and Ramesh, Aditya and Goh, Gabriel and Agarwal, Sandhini and Sastry, Girish and Askell, Amanda and Mishkin, Pamela and Clark, Jack and others},
  booktitle={International conference on machine learning},
  pages={8748--8763},
  year={2021},
  organization={PMLR}
}

@article{li2023blip,
  title={Blip-2: Bootstrapping language-image pre-training with frozen image encoders and large language models},
  author={Li, Junnan and Li, Dongxu and Savarese, Silvio and Hoi, Steven},
  journal=icml,
  year={2023}
}

@inproceedings{wang2020minegan,
  title={Minegan: effective knowledge transfer from gans to target domains with few images},
  author={Wang, Yaxing and Gonzalez-Garcia, Abel and Berga, David and Herranz, Luis and Khan, Fahad Shahbaz and Weijer, Joost van de},
  booktitle={Proceedings of the IEEE/CVF conference on computer vision and pattern recognition},
  pages={9332--9341},
  year={2020}
}

@article{xing2024survey_vdm,
  title={A survey on video diffusion models},
  author={Xing, Zhen and Feng, Qijun and Chen, Haoran and Dai, Qi and Hu, Han and Xu, Hang and Wu, Zuxuan and Jiang, Yu-Gang},
  journal={ACM Computing Surveys},
  volume={57},
  number={2},
  pages={1--42},
  year={2024},
  publisher={ACM New York, NY}
}

@article{zhang2023text2image_survey,
  title={Text-to-image diffusion models in generative ai: A survey},
  author={Zhang, Chenshuang and Zhang, Chaoning and Zhang, Mengchun and Kweon, In So},
  journal={arXiv preprint arXiv:2303.07909},
  year={2023}
}

@article{ma2023overview_video_coding,
  title={Overview of intelligent video coding: from model-based to learning-based approaches},
  author={Ma, Siwei and Gao, Junlong and Wang, Ruofan and Chang, Jianhui and Mao, Qi and Huang, Zhimeng and Jia, Chuanmin},
  journal={Visual Intelligence},
  volume={1},
  number={1},
  pages={15},
  year={2023},
  publisher={Springer}
}

@article{tu2024overview_large_AI,
  title={An overview of large AI models and their applications},
  author={Tu, Xiaoguang and He, Zhi and Huang, Yi and Zhang, Zhi-Hao and Yang, Ming and Zhao, Jian},
  journal={Visual Intelligence},
  volume={2},
  number={1},
  pages={1--22},
  year={2024},
  publisher={Springer}
}

@inproceedings{zhang2024ssr_encoder,
  title={Ssr-encoder: Encoding selective subject representation for subject-driven generation},
  author={Zhang, Yuxuan and Song, Yiren and Liu, Jiaming and Wang, Rui and Yu, Jinpeng and Tang, Hao and Li, Huaxia and Tang, Xu and Hu, Yao and Pan, Han and others},
  booktitle={Proceedings of the IEEE/CVF Conference on Computer Vision and Pattern Recognition},
  pages={8069--8078},
  year={2024}
}

@inproceedings{wu2024infinite_id,
  title={Infinite-ID: Identity-preserved Personalization via ID-semantics Decoupling Paradigm},
  author={Wu, Yi and Li, Ziqiang and Zheng, Heliang and Wang, Chaoyue and Li, Bin},
  booktitle={European Conference on Computer Vision},
  pages={279--296},
  year={2024},
  organization={Springer}
}

@inproceedings{cui2024idadapter,
  title={Idadapter: Learning mixed features for tuning-free personalization of text-to-image models},
  author={Cui, Siying and Guo, Jia and An, Xiang and Deng, Jiankang and Zhao, Yongle and Wei, Xinyu and Feng, Ziyong},
  booktitle={Proceedings of the IEEE/CVF Conference on Computer Vision and Pattern Recognition},
  pages={950--959},
  year={2024}
}

@article{guo2024pulid,
  title={Pulid: Pure and lightning id customization via contrastive alignment},
  author={Guo, Zinan and Wu, Yanze and Zhuowei, Chen and Zhang, Peng and He, Qian and others},
  journal={Advances in neural information processing systems},
  volume={37},
  pages={36777--36804},
  year={2024}
}

@article{achlioptas2023stellar,
  title={Stellar: systematic evaluation of human-centric personalized text-to-image methods},
  author={Achlioptas, Panos and Benetatos, Alexandros and Fostiropoulos, Iordanis and Skourtis, Dimitris},
  journal={arXiv preprint arXiv:2312.06116},
  year={2023}
}

@article{xiang2023closer_look,
  title={A closer look at parameter-efficient tuning in diffusion models},
  author={Xiang, Chendong and Bao, Fan and Li, Chongxuan and Su, Hang and Zhu, Jun},
  journal={arXiv preprint arXiv:2303.18181},
  year={2023}
}

@article{chen2023suti,
  title={Subject-driven Text-to-Image Generation via Apprenticeship Learning},
  author={Chen, Wenhu and Hu, Hexiang and Li, Yandong and Rui, Nataniel and Jia, Xuhui and Chang, Ming-Wei and Cohen, William W},
  journal=nips,
  year={2023}
}

@article{shi2023instantbooth,
  title={InstantBooth: Personalized Text-to-Image Generation without Test-Time Finetuning},
  author={Shi, Jing and Xiong, Wei and Lin, Zhe and Jung, Hyun Joon},
  journal=cvpr,
  year={2024}
}

@article{zhang2022sine,
  title={SINE: SINgle Image Editing with Text-to-Image Diffusion Models},
  author={Zhang, Zhixing and Han, Ligong and Ghosh, Arnab and Metaxas, Dimitris and Ren, Jian},
  journal=cvpr,
  year={2023}
}

@article{textual_inversion,
  title={An image is worth one word: Personalizing text-to-image generation using textual inversion},
  author={Gal, Rinon and Alaluf, Yuval and Atzmon, Yuval and Patashnik, Or and Bermano, Amit H and Chechik, Gal and Cohen-Or, Daniel},
  journal=iclr,
  year={2023}
}

@article{kumari2023customdiffusion,
  title={Multi-Concept Customization of Text-to-Image Diffusion},
  author={Kumari, Nupur and Zhang, Bingliang and Zhang, Richard and Shechtman, Eli and Zhu, Jun-Yan},
  journal=cvpr,
  year={2023}
}

@article{gal2023e4t,
  title={Designing an Encoder for Fast Personalization of Text-to-Image Models},
  author={Gal, Rinon and Arar, Moab and Atzmon, Yuval and Bermano, Amit H and Chechik, Gal and Cohen-Or, Daniel},
  journal={arXiv preprint arXiv:2302.12228},
  year={2023}
}

@article{dong2022dreamartist,
  title={Dreamartist: Towards controllable one-shot text-to-image generation via contrastive prompt-tuning},
  author={Dong, Ziyi and Wei, Pengxu and Lin, Liang},
  journal={arXiv preprint arXiv:2211.11337},
  year={2022}
}

@article{Cones2023,
  author = {Liu, Zhiheng and Feng, Ruili and Zhu, Kai and Zhang, Yifei and Zheng, Kecheng and Liu, Yu and Zhao, Deli and Zhou, Jingren and Cao, Yang},
  title = {Cones: Concept Neurons in Diffusion Models for Customized Generation},
  journal = icml,
  year = {2023},
}

@article{patel2024lambda_eclipse,
  title={lambda-ECLIPSE: Multi-Concept Personalized Text-to-Image Diffusion Models by Leveraging CLIP Latent Space},
  author={Patel, Maitreya and Jung, Sangmin and Baral, Chitta and Yang, Yezhou},
  journal={arXiv preprint arXiv:2402.05195},
  year={2024}
}

@article{shah2023ziplora,
  title={ZipLoRA: Any Subject in Any Style by Effectively Merging LoRAs},
  author={Shah, Viraj and Ruiz, Nataniel and Cole, Forrester and Lu, Erika and Lazebnik, Svetlana and Li, Yuanzhen and Jampani, Varun},
  journal={arXiv preprint arXiv:2311.13600},
  year={2023}
}

@article{sun2023generative_emu2,
  title={Generative multimodal models are in-context learners},
  author={Sun, Quan and Cui, Yufeng and Zhang, Xiaosong and Zhang, Fan and Yu, Qiying and Luo, Zhengxiong and Wang, Yueze and Rao, Yongming and Liu, Jingjing and Huang, Tiejun and others},
  journal=cvpr,
  year={2024}
}

@article{smith2023continual,
  title={Continual diffusion: Continual customization of text-to-image diffusion with c-lora},
  author={Smith, James Seale and Hsu, Yen-Chang and Zhang, Lingyu and Hua, Ting and Kira, Zsolt and Shen, Yilin and Jin, Hongxia},
  journal={arXiv preprint arXiv:2304.06027},
  year={2023}
}

@article{Wei2023ELITEEV,
  title={ELITE: Encoding Visual Concepts into Textual Embeddings for Customized Text-to-Image Generation},
  author={Yuxiang Wei and Yabo Zhang and Zhilong Ji and Jinfeng Bai and Lei Zhang and Wangmeng Zuo},
  journal={2023 IEEE/CVF International Conference on Computer Vision (ICCV)},
  year={2023},
  pages={15897-15907},
  url={https://api.semanticscholar.org/CorpusID:257219968}
}

@article{ruiz2023hyperdreambooth,
  title={Hyperdreambooth: Hypernetworks for fast personalization of text-to-image models},
  author={Ruiz, Nataniel and Li, Yuanzhen and Jampani, Varun and Wei, Wei and Hou, Tingbo and Pritch, Yael and Wadhwa, Neal and Rubinstein, Michael and Aberman, Kfir},
  journal={arXiv preprint arXiv:2307.06949},
  year={2023}
}

@article{jia2023taming,
  title={Taming encoder for zero fine-tuning image customization with text-to-image diffusion models},
  author={Jia, Xuhui and Zhao, Yang and Chan, Kelvin CK and Li, Yandong and Zhang, Han and Gong, Boqing and Hou, Tingbo and Wang, Huisheng and Su, Yu-Chuan},
  journal={arXiv preprint arXiv:2304.02642},
  year={2023}
}

@article{wang2023prolificdreamer,
  title={Prolificdreamer: High-fidelity and diverse text-to-3d generation with variational score distillation},
  author={Wang, Zhengyi and Lu, Cheng and Wang, Yikai and Bao, Fan and Li, Chongxuan and Su, Hang and Zhu, Jun},
  journal={Advances in Neural Information Processing Systems},
  volume={36},
  pages={8406--8441},
  year={2023}
}

@inproceedings{gal2024lcm_lookahead,
  title={Lcm-lookahead for encoder-based text-to-image personalization},
  author={Gal, Rinon and Lichter, Or and Richardson, Elad and Patashnik, Or and Bermano, Amit H and Chechik, Gal and Cohen-Or, Daniel},
  booktitle={European Conference on Computer Vision},
  pages={322--340},
  year={2024},
  organization={Springer}
}

@article{simsar2024loraclr,
  title={LoRACLR: Contrastive Adaptation for Customization of Diffusion Models},
  author={Simsar, Enis and Hofmann, Thomas and Tombari, Federico and Yanardag, Pinar},
  journal=cvpr,
  year={2025}
}

@inproceedings{chen2022re,
  title={Re-Imagen: Retrieval-Augmented Text-to-Image Generator},
  author={Chen, Wenhu and Hu, Hexiang and Saharia, Chitwan and Cohen, William W},
  booktitle={The Eleventh International Conference on Learning Representations},
  year={2022}
}

@article{jiang2025infiniteyou,
  title={InfiniteYou: Flexible Photo Recrafting While Preserving Your Identity},
  author={Jiang, Liming and Yan, Qing and Jia, Yumin and Liu, Zichuan and Kang, Hao and Lu, Xin},
  journal=iccv,
  year={2025}
}

@article{pan2023kosmos,
  title={Kosmos-g: Generating images in context with multimodal large language models},
  author={Pan, Xichen and Dong, Li and Huang, Shaohan and Peng, Zhiliang and Chen, Wenhu and Wei, Furu},
  journal=iclr,
  year={2024}
}

@article{ma2023unified,
  title={Unified multi-modal latent diffusion for joint subject and text conditional image generation},
  author={Ma, Yiyang and Yang, Huan and Wang, Wenjing and Fu, Jianlong and Liu, Jiaying},
  journal={arXiv preprint arXiv:2303.09319},
  year={2023}
}

@article{ma2023subject_diffusion,
  title={Subject-diffusion: Open domain personalized text-to-image generation without test-time fine-tuning},
  author={Ma, Jian and Liang, Junhao and Chen, Chen and Lu, Haonan},
  journal=siggraph,
  year={2024}
}

@article{alaluf2023neural_neti,
  title={A Neural Space-Time Representation for Text-to-Image Personalization},
  author={Alaluf, Yuval and Richardson, Elad and Metzer, Gal and Cohen-Or, Daniel},
  journal={ACM Transactions on Graphics (TOG)},
  volume={42},
  number={6},
  pages={1--10},
  year={2023},
  publisher={ACM New York, NY, USA}
}

@inproceedings{liu2023customizable_cones2,
  title={Customizable Image Synthesis with Multiple Subjects},
  author={Liu, Zhiheng and Zhang, Yifei and Shen, Yujun and Zheng, Kecheng and Zhu, Kai and Feng, Ruili and Liu, Yu and Zhao, Deli and Zhou, Jingren and Cao, Yang},
  booktitle={Thirty-seventh Conference on Neural Information Processing Systems},
  year={2023}
}

@article{han2023svdiff,
      title={SVDiff: Compact Parameter Space for Diffusion Fine-Tuning}, 
      author={Ligong Han and Yinxiao Li and Han Zhang and Peyman Milanfar and Dimitris Metaxas and Feng Yang},
      year={2023},
      journal=iccv,
}

@article{voynov2023ETI,
  title={$P+$: Extended Textual Conditioning in Text-to-Image Generation},
  author={Voynov, Andrey and Chu, Qinghao and Cohen-Or, Daniel and Aberman, Kfir},
  journal={arXiv preprint arXiv:2303.09522},
  year={2023}
}

@inproceedings{ronneberger2015unet,
  title={U-net: Convolutional networks for biomedical image segmentation},
  author={Ronneberger, Olaf and Fischer, Philipp and Brox, Thomas},
  booktitle={Medical Image Computing and Computer-Assisted Intervention--MICCAI 2015: 18th International Conference, Munich, Germany, October 5-9, 2015, Proceedings, Part III 18},
  pages={234--241},
  year={2015},
  organization={Springer}
}

@article{ho2020ddpm,
  title={Denoising diffusion probabilistic models},
  author={Ho, Jonathan and Jain, Ajay and Abbeel, Pieter},
  journal={Advances in Neural Information Processing Systems},
  volume={33},
  pages={6840--6851},
  year={2020}
}

@inproceedings{caron2021dino,
  title={Emerging properties in self-supervised vision transformers},
  author={Caron, Mathilde and Touvron, Hugo and Misra, Ishan and J{\'e}gou, Herv{\'e} and Mairal, Julien and Bojanowski, Piotr and Joulin, Armand},
  booktitle={Proceedings of the IEEE/CVF international conference on computer vision},
  pages={9650--9660},
  year={2021}
}

@article{tang2023iterinv,
  title={IterInv: Iterative Inversion for Pixel-Level T2I Models},
  author={Tang, Chuanming and Wang, Kai and van de Weijer, Joost},
  journal={Neurips 2023 workshop on Diffusion Models},
  year={2023}
}

@article{avrahami2023breakascene,
  title={Break-A-Scene: Extracting Multiple Concepts from a Single Image},
  author={Avrahami, Omri and Aberman, Kfir and Fried, Ohad and Cohen-Or, Daniel and Lischinski, Dani},
  journal={SIGGRAPH Asia 2023},
  year={2023}
}

@article{chen2023disenbooth,
  title={Disenbooth: Identity-preserving disentangled tuning for subject-driven text-to-image generation},
  author={Chen, Hong and Zhang, Yipeng and Wu, Simin and Wang, Xin and Duan, Xuguang and Zhou, Yuwei and Zhu, Wenwu},
  journal=iclr,
  year={2024}
}

@article{zhao2025catversion,
  title={Catversion: Concatenating embeddings for diffusion-based text-to-image personalization},
  author={Zhao, Ruoyu and Zhu, Mingrui and Dong, Shiyin and Cheng, De and Wang, Nannan and Gao, Xinbo},
  journal={IEEE Transactions on Circuits and Systems for Video Technology},
  year={2025},
  publisher={IEEE}
}

@inproceedings{arar2023domain_agnostic,
  title={Domain-agnostic tuning-encoder for fast personalization of text-to-image models},
  author={Arar, Moab and Gal, Rinon and Atzmon, Yuval and Chechik, Gal and Cohen-Or, Daniel and Shamir, Ariel and H. Bermano, Amit},
  booktitle={SIGGRAPH Asia 2023 Conference Papers},
  pages={1--10},
  year={2023}
}

@article{arkhipkin2023kandinsky,
  title={Kandinsky 3.0 technical report},
  author={Arkhipkin, Vladimir and Filatov, Andrei and Vasilev, Viacheslav and Maltseva, Anastasia and Azizov, Said and Pavlov, Igor and Agafonova, Julia and Kuznetsov, Andrey and Dimitrov, Denis},
  journal={arXiv preprint arXiv:2312.03511},
  year={2023}
}

@article{li2023photomaker,
  title={Photomaker: Customizing realistic human photos via stacked id embedding},
  author={Li, Zhen and Cao, Mingdeng and Wang, Xintao and Qi, Zhongang and Cheng, Ming-Ming and Shan, Ying},
  journal=cvpr,
  year={2024}
}

@article{gu2024mixofshow,
  title={Mix-of-show: Decentralized low-rank adaptation for multi-concept customization of diffusion models},
  author={Gu, Yuchao and Wang, Xintao and Wu, Jay Zhangjie and Shi, Yujun and Chen, Yunpeng and Fan, Zihan and Xiao, Wuyou and Zhao, Rui and Chang, Shuning and Wu, Weijia and others},
  journal={Advances in Neural Information Processing Systems},
  volume={36},
  year={2024}
}

@inproceedings{zhang2023controlnet,
  title={Adding conditional control to text-to-image diffusion models},
  author={Zhang, Lvmin and Rao, Anyi and Agrawala, Maneesh},
  booktitle={Proceedings of the IEEE/CVF International Conference on Computer Vision},
  pages={3836--3847},
  year={2023}
}

@article{huang2024consistentid,
  title={Consistentid: Portrait generation with multimodal fine-grained identity preserving},
  author={Huang, Jiehui and Dong, Xiao and Song, Wenhui and Li, Hanhui and Zhou, Jun and Cheng, Yuhao and Liao, Shutao and Chen, Long and Yan, Yiqiang and Liao, Shengcai and others},
  journal={arXiv preprint arXiv:2404.16771},
  year={2024}
}

@misc{oquab2023dinov2,
  title={DINOv2: Learning Robust Visual Features without Supervision},
  author={Oquab, Maxime and Darcet, Timothée and Moutakanni, Theo and Vo, Huy V. and Szafraniec, Marc and Khalidov, Vasil and Fernandez, Pierre and Haziza, Daniel and Massa, Francisco and El-Nouby, Alaaeldin and Howes, Russell and Huang, Po-Yao and Xu, Hu and Sharma, Vasu and Li, Shang-Wen and Galuba, Wojciech and Rabbat, Mike and Assran, Mido and Ballas, Nicolas and Synnaeve, Gabriel and Misra, Ishan and Jegou, Herve and Mairal, Julien and Labatut, Patrick and Joulin, Armand and Bojanowski, Piotr},
  journal={arXiv:2304.07193},
  year={2023}
}

@inproceedings{ruiz2023dreambooth,
  title={Dreambooth: Fine tuning text-to-image diffusion models for subject-driven generation},
  author={Ruiz, Nataniel and Li, Yuanzhen and Jampani, Varun and Pritch, Yael and Rubinstein, Michael and Aberman, Kfir},
  booktitle={Proceedings of the IEEE/CVF Conference on Computer Vision and Pattern Recognition},
  pages={22500--22510},
  year={2023}
}

@inproceedings{zeng2024jedi,
  title={Jedi: Joint-image diffusion models for finetuning-free personalized text-to-image generation},
  author={Zeng, Yu and Patel, Vishal M and Wang, Haochen and Huang, Xun and Wang, Ting-Chun and Liu, Ming-Yu and Balaji, Yogesh},
  booktitle={Proceedings of the IEEE/CVF Conference on Computer Vision and Pattern Recognition},
  pages={6786--6795},
  year={2024}
}

@inproceedings{tewel2023keylocked_perfusion,
  title={Key-locked rank one editing for text-to-image personalization},
  author={Tewel, Yoad and Gal, Rinon and Chechik, Gal and Atzmon, Yuval},
  booktitle={ACM SIGGRAPH 2023 Conference Proceedings},
  pages={1--11},
  year={2023}
}

@article{zhang2023prospect,
  title={ProSpect: Expanded Conditioning for the Personalization of Attribute-aware Image Generation},
  author={Zhang, Yuxin and Dong, Weiming and Tang, Fan and Huang, Nisha and Huang, Haibin and Ma, Chongyang and Lee, Tong-Yee and Deussen, Oliver and Xu, Changsheng},
  journal={SIGGRAPH Asia 2023},
  year={2023}
}

@article{salimans2022progressive,
  title={Progressive distillation for fast sampling of diffusion models},
  author={Salimans, Tim and Ho, Jonathan},
  journal=iclr,
  year={2022}
}

@article{mou2023t2i,
  title={T2i-adapter: Learning adapters to dig out more controllable ability for text-to-image diffusion models},
  author={Mou, Chong and Wang, Xintao and Xie, Liangbin and Zhang, Jian and Qi, Zhongang and Shan, Ying and Qie, Xiaohu},
  journal=aaai,
  year={2024}
}

@article{poole2023dreamfusion,
  author = {Poole, Ben and Jain, Ajay and Barron, Jonathan T. and Mildenhall, Ben},
  title = {{DreamFusion: Text-to-3D using 2D Diffusion}},
  journal = ICLR,
  year = {2023},
}

@article{podell2023sdxl,
  title={Sdxl: Improving latent diffusion models for high-resolution image synthesis},
  author={Podell, Dustin and English, Zion and Lacey, Kyle and Blattmann, Andreas and Dockhorn, Tim and M{\"u}ller, Jonas and Penna, Joe and Rombach, Robin},
  journal={arXiv preprint arXiv:2307.01952},
  year={2023}
}

@article{nguyen2023swiftbrush,
  title={SwiftBrush: One-Step Text-to-Image Diffusion Model with Variational Score Distillation},
  author={Nguyen, Thuan Hoang and Tran, Anh},
  journal=cvpr,
  year={2024}
}

@article{dao2024swiftbrushv2,
  title={SwiftBrush v2: Make Your One-step Diffusion Model Better Than Its Teacher},
  author={Dao, Trung and Nguyen, Thuan Hoang and Le, Thanh and Vu, Duc and Nguyen, Khoi and Pham, Cuong and Tran, Anh},
  journal=eccv,
  year={2024}
}

@article{yin2024dmd2,
  title={Improved Distribution Matching Distillation for Fast Image Synthesis},
  author={Yin, Tianwei and Gharbi, Micha{\"e}l and Park, Taesung and Zhang, Richard and Shechtman, Eli and Durand, Fredo and Freeman, William T},
  journal=nips,
  year={2024}
}

@article{luo2023lcm_lora,
  title={Lcm-lora: A universal stable-diffusion acceleration module},
  author={Luo, Simian and Tan, Yiqin and Patil, Suraj and Gu, Daniel and von Platen, Patrick and Passos, Apolin{\'a}rio and Huang, Longbo and Li, Jian and Zhao, Hang},
  journal={arXiv preprint arXiv:2311.05556},
  year={2023}
}

@inproceedings{song2023consistency,
  title={Consistency Models},
  author={Song, Yang and Dhariwal, Prafulla and Chen, Mark and Sutskever, Ilya},
  booktitle=icml,
  pages={32211--32252},
  year={2023},
  organization={PMLR}
}

@article{luo2023latent,
  title={Latent Consistency Models: Synthesizing High-Resolution Images with Few-Step Inference}, 
  author={Simian Luo and Yiqin Tan and Longbo Huang and Jian Li and Hang Zhao},
  year={2023},
  journal={arXiv preprint arXiv:2310.04378},
}

@article{sauer2023adversarial,
  title={Adversarial Diffusion Distillation},
  author={Sauer, Axel and Lorenz, Dominik and Blattmann, Andreas and Rombach, Robin},
  journal=eccv,
  year={2024}
}

@inproceedings{chan2022eg3d,
  title={Efficient geometry-aware 3d generative adversarial networks},
  author={Chan, Eric R and Lin, Connor Z and Chan, Matthew A and Nagano, Koki and Pan, Boxiao and De Mello, Shalini and Gallo, Orazio and Guibas, Leonidas J and Tremblay, Jonathan and Khamis, Sameh and others},
  booktitle={Proceedings of the IEEE/CVF conference on computer vision and pattern recognition},
  pages={16123--16133},
  year={2022}
}

@article{lu2022dpm,
  title={Dpm-solver: A fast ode solver for diffusion probabilistic model sampling in around 10 steps},
  author={Lu, Cheng and Zhou, Yuhao and Bao, Fan and Chen, Jianfei and Li, Chongxuan and Zhu, Jun},
  journal={Advances in Neural Information Processing Systems},
  volume={35},
  pages={5775--5787},
  year={2022}
}

@inproceedings{
song2021ddim,
title={Denoising Diffusion Implicit Models},
author={Jiaming Song and Chenlin Meng and Stefano Ermon},
booktitle={International Conference on Learning Representations},
year={2021},
url={https://openreview.net/forum?id=St1giarCHLP}
}

@article{zheng2024trajectory_tcd,
  title={Trajectory consistency distillation},
  author={Zheng, Jianbin and Hu, Minghui and Fan, Zhongyi and Wang, Chaoyue and Ding, Changxing and Tao, Dacheng and Cham, Tat-Jen},
  journal={arXiv preprint arXiv:2402.19159},
  year={2024}
}

@article{agarwal2023image_matte,
  title={An image is worth multiple words: Multi-attribute inversion for constrained text-to-image synthesis},
  author={Agarwal, Aishwarya and Karanam, Srikrishna and Shukla, Tripti and Srinivasan, Balaji Vasan},
  journal=icml,
  year={2024}
}

@inproceedings{butt2025colorpeel,
  title={ColorPeel: Color Prompt Learning with Diffusion Models via Color and Shape Disentanglement},
  author={Butt, Muhammad Atif and Wang, Kai and Vazquez-Corral, Javier and van de Weijer, Joost},
  booktitle=ECCV,
  year={2024},
}

@article{hu2024token_merging_tome,
  title={Token Merging for Training-Free Semantic Binding in Text-to-Image Synthesis},
  author={Hu, Taihang and Li, Linxuan and van de Weijer, Joost and Gao, Hongcheng and Shahbaz Khan, Fahad and Yang, Jian and Cheng, Ming-Ming and Wang, Kai and Wang, Yaxing},
  journal={Advances in Neural Information Processing Systems},
  volume={37},
  pages={137646--137672},
  year={2024}
}

@article{kai2023DPL,
  author = {Kai Wang and Fei Yang and Shiqi Yang and Muhammad Atif Butt and Joost van de Weijer},
  title = {Dynamic Prompt Learning: Addressing Cross-Attention Leakage for
  Text-Based Image Editing},
  journal = nips,
  year = {2023},
}

@article{liu2025onepromptonestory,
  title={One-Prompt-One-Story: Free-Lunch Consistent Text-to-Image Generation Using a Single Prompt},
  author={Liu, Tao and Wang, Kai and Li, Senmao and van de Weijer, Joost and Khan, Fahad Shahbaz and Yang, Shiqi and Wang, Yaxing and Yang, Jian and Cheng, Ming-Ming},
  journal=cvpr,
  year={2025}
}

@inproceedings{wang2024mcti,
  title={Multi-Class Textual-Inversion Secretly Yields a Semantic-Agnostic Classifier},
  author={Wang, Kai and Yang, Fei and Raducanu, Bogdan and van de Weijer, Joost},
  booktitle=wacv,
  year={2025}
}

@article{tang2024locinv,
  title={Locinv: localization-aware inversion for text-guided image editing},
  author={Tang, Chuanming and Wang, Kai and Yang, Fei and van de Weijer, Joost},
  journal={arXiv preprint arXiv:2405.01496},
  year={2024}
}

@inproceedings{he2024multiscale,
  title={Multiscale Sliced Wasserstein Distances as Perceptual Color Difference Measures},
  author={He, Jiaqi and Wang, Zhihua and Wang, Leon and Liu, Tsein-I and Fang, Yuming and Sun, Qilin and Ma, Kede},
  booktitle={European Conference on Computer Vision},
  pages={425--442},
  year={2024},
  organization={Springer}
}

@inproceedings{kumari2022ensembling,
  title={Ensembling off-the-shelf models for gan training},
  author={Kumari, Nupur and Zhang, Richard and Shechtman, Eli and Zhu, Jun-Yan},
  booktitle={Proceedings of the IEEE/CVF conference on computer vision and pattern recognition},
  pages={10651--10662},
  year={2022}
}

@inproceedings{tan2025ominicontrol,
  title={OminiControl: Minimal and Universal Control for Diffusion Transformer},
  author={Tan, Zhenxiong and Liu, Songhua and Yang, Xingyi and Xue, Qiaochu and Wang, Xinchao},
  booktitle={Proceedings of the IEEE/CVF International Conference on Computer Vision},
  year={2025}
}

@article{mou2025dreamo,
  title={Dreamo: A unified framework for image customization},
  author={Mou, Chong and Wu, Yanze and Wu, Wenxu and Guo, Zinan and Zhang, Pengze and Cheng, Yufeng and Luo, Yiming and Ding, Fei and Zhang, Shiwen and Li, Xinghui and others},
  journal={SIGGRAPH Asia},
  year={2025}
}

@article{chen2025xverse,
  title={XVerse: Consistent Multi-Subject Control of Identity and Semantic Attributes via DiT Modulation},
  author={Chen, Bowen and Zhao, Mengyi and Sun, Haomiao and Chen, Li and Wang, Xu and Du, Kang and Wu, Xinglong},
  journal={arXiv preprint arXiv:2506.21416},
  year={2025}
}

@article{wu2025less,
  title={Less-to-More Generalization: Unlocking More Controllability by In-Context Generation},
  author={Wu, Shaojin and Huang, Mengqi and Wu, Wenxu and Cheng, Yufeng and Ding, Fei and He, Qian},
  journal={arXiv preprint arXiv:2504.02160},
  year={2025}
}

@article{deng2025bagel,
  title   = {Emerging Properties in Unified Multimodal Pretraining},
  author  = {Deng, Chaorui and Zhu, Deyao and Li, Kunchang and Gou, Chenhui and Li, Feng and Wang, Zeyu and Zhong, Shu and Yu, Weihao and Nie, Xiaonan and Song, Ziang and Shi, Guang and Fan, Haoqi},
  journal = {arXiv preprint arXiv:2505.14683},
  year    = {2025}
}

@article{achiam2023gpt,
  title={Gpt-4 technical report},
  author={Achiam, Josh and Adler, Steven and Agarwal, Sandhini and Ahmad, Lama and Akkaya, Ilge and Aleman, Florencia Leoni and Almeida, Diogo and Altenschmidt, Janko and Altman, Sam and Anadkat, Shyamal and others},
  journal={arXiv preprint arXiv:2303.08774},
  year={2023}
}

@article{wu2025omnigen2,
  title={OmniGen2: Exploration to Advanced Multimodal Generation},
  author={Wu, Chenyuan and Zheng, Pengfei and Yan, Ruiran and Xiao, Shitao and Luo, Xin and Wang, Yueze and Li, Wanli and Jiang, Xiyan and Liu, Yexin and Zhou, Junjie and others},
  journal={arXiv preprint arXiv:2506.18871},
  year={2025}
}

@article{kumari2025generating,
  title={Generating multi-image synthetic data for text-to-image customization},
  author={Kumari, Nupur and Yin, Xi and Zhu, Jun-Yan and Misra, Ishan and Azadi, Samaneh},
  journal={arXiv preprint arXiv:2502.01720},
  year={2025}
}

@article{sabour2025align,
  title={Align Your Flow: Scaling Continuous-Time Flow Map Distillation},
  author={Sabour, Amirmojtaba and Fidler, Sanja and Kreis, Karsten},
  journal={arXiv preprint arXiv:2506.14603},
  year={2025}
}

@InProceedings{Luo_2025_ICCV,
    author    = {Luo, Yihong and Hu, Tianyang and Song, Yifan and Sun, Jiacheng and Li, Zhenguo and Tang, Jing},
    title     = {Adding Additional Control to One-Step Diffusion with Joint Distribution Matching},
    booktitle = {Proceedings of the IEEE/CVF International Conference on Computer Vision (ICCV)},
    month     = {October},
    year      = {2025},
    pages     = {4009-4018}
}
